\documentclass{article}
\usepackage{setspace}
\PassOptionsToPackage{numbers}{natbib}
\usepackage[final]{neurips_2022}

\usepackage{microtype}
\usepackage{graphicx}
\usepackage{booktabs} 

\usepackage{caption}

\usepackage[position=bottom]{subfig}
\usepackage{arydshln}
\usepackage[export]{adjustbox}
\usepackage{multirow}
\usepackage{enumitem} 
\usepackage{placeins} 
\usepackage{pifont}
\newcommand{\cmark}{\ding{51}}%
\newcommand{\xmark}{\ding{55}}%
\usepackage{xcolor}
\definecolor{myblue}{rgb}{0.0,0.08,1.0}
\usepackage{bm}

\usepackage{hyperref}
\hypersetup{
    colorlinks,
    citecolor=myblue,
    filecolor=myblue,
    linkcolor=myblue,
    urlcolor=myblue
}

\usepackage{amsmath}
\usepackage{amssymb}
\usepackage{mathtools}
\usepackage{amsthm}

\usepackage[capitalize,noabbrev]{cleveref}

\usepackage{comment}
\newcommand{\todo}[1]{\textcolor{red}{#1}}

\begin{document}

\title{
Regression Transformer: Concurrent sequence regression and generation for molecular language modeling
}




\author{%
  Jannis Born\\
  IBM Research Europe, Zurich, Switzerland \\
  Department of Biosystem Science and Engineering, ETH Zurich, Basel, Switzerland \\
  \texttt{jab@zurich.ibm.com} \\
  \And
  Matteo Manica \\
  IBM Research Europe, Zurich, Switzerland \\
  \texttt{tte@zurich.ibm.com} \\
}









\maketitle
\begin{abstract}
Despite significant progress of generative models in the natural sciences, their controllability remains challenging. 
One fundamentally missing aspect of molecular or protein generative models is an inductive bias that can reflect continuous properties of interest. 
%
To that end, we propose the Regression Transformer (RT), a novel method that abstracts regression as a conditional sequence modeling problem. 
This introduces a new paradigm of multitask language models which 
seamlessly bridge sequence regression and conditional sequence generation.

We thoroughly demonstrate that, despite using a nominal-scale training objective, the RT matches or surpasses the performance of conventional regression models in property prediction tasks of small molecules, proteins and chemical reactions.

Critically, priming the same model with continuous properties yields a highly competitive conditional generative model that outperforms specialized approaches in a substructure-constrained, property-driven molecule generation benchmark.
Our dichotomous approach is facilitated by a novel, alternating training scheme that enables the model to decorate seed sequences by desired properties, e.g., to optimize reaction yield.

In sum, the RT is the first report of a multitask model that concurrently excels at predictive and generative tasks in biochemistry.
This finds particular application in property-driven, local exploration of the chemical or protein space and could pave the road toward foundation models in material design.


The code to reproduce all experiments of the paper is available at:~
\url{https://github.com/IBM/regression-transformer}
\end{abstract}

\section{Introduction}
Transformers~\citep{vaswani2017attention} are now ubiquitous in natural language processing (NLP) and have also enjoyed large success in molecular~\citep{schwaller2019molecular,schwaller2021mapping,schwaller2021extraction}
and protein language modeling~\citep{rives2021biological,jumper2021highly}.
The invention of Transformers was in alignment with the steady decline of inductive biases in ML, a trend that started with the rise of deep learning:
CNNs outperformed traditional feature descriptors in object recognition~\citep{krizhevsky2012imagenet}, self-attention generalized dense layers to learn sample-dependent instead of static affine transformations~\citep{luong2015effective} and Transformers exploited self-attention 
to supersede RNNs as
the de-facto standard in NLP.
The success of vision transformers has questioned the need for translation equivariance 
in image processing~\citep{ramachandran2019stand} and now, even frozen Transformers pretrained on text achieve SOTA results in object detection and protein classification~\citep{lu2022frozen}.
Given that Transformers are today's most generic model\footnote{graph neural networks with multihead attention as neighborhood aggregation on complete graphs.
},
it is not surprising that attempts have been made to abstract entire domains like RL to sequence modeling in order to leverage Transformers~\citep{chen2021decision}.
%
\begin{figure*}[!htb]
\centering
\includegraphics[width=1\linewidth]{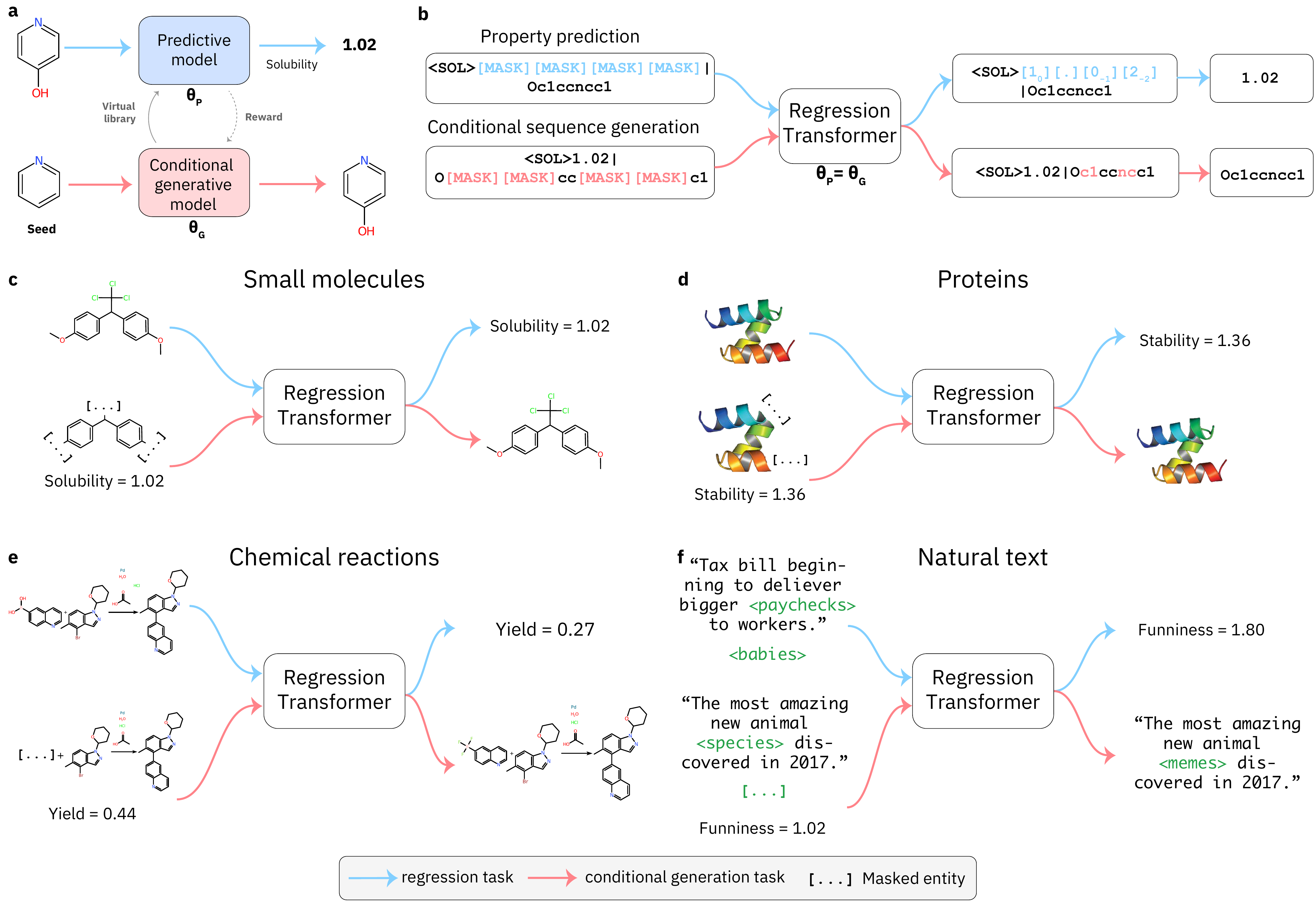}
\caption{
\begin{small}
\textbf{Overview of Regression Transformer (RT).}
The RT is a multitask language model designed to handle combinations of text and numbers.
\textbf{a)} Traditional approach in generative chemistry: property predictors and generative models are trained independently from another.
\textbf{b)} Our approach:  Training the RT yields a dichotomous model that seamlessly switches between property prediction and conditional text generation.
The model's task is to fill the content behind the~\texttt{[MASK]} tokens.
Depending on the mask location, the same model either predicts numerical tokens given textual tokens, thus performing a regression task \textit{(blue stream, top)}; or predicts textual tokens given both numerical and textual tokens, thus performing a property-driven conditional generation \textit{(yellow stream, bottom)}.
\textbf{c)} - \textbf{f)}:
This novel formulation finds application across a wide range of domains. 
We demonstrate the flexibility of the RT in predictive and generative tasks in modeling small molecules, proteins, chemical reactions and even natural text.
\end{small}
}
\label{fig:graphicalabstract}
\end{figure*}

A provocative next step toward reducing inductive biases might be to refrain from explicitly modeling target variables as functions of input variables.
Instead of following this discriminative modelling approach when tuning task-specific language heads in Transformers, learning the joint distribution over input and target variables could effectively further blur the lines between predictive and conditional generative models.
The feasibility of such approach can be assessed via permutation language modeling (PLM), an extension of masked-language-modeling to autoregressive models~\citep{yang2019xlnet}.
Such dichotomous models (that concurrently excel at regression and conditional sequence generation) are
beyond applications in NLP
of special interest for in chemical and material design.
Molecules are often labelled with continuous properties (e.g., drug efficacy or protein solubility) and design tasks are intertwined with bio- or physicochemical properties.
But despite the rise of deep generative models in molecular~\cite{elton2019deep, chen2021deep} and protein design~\cite{wu2021protein,madani2020progen}, current approaches still develop property predictors and generative models independently.
Transformer-based architectures have been used widely on chemical tasks but either focused on property prediction~\citep{wang2019smiles,kim2021merged} or on conditional molecular design~\citep{irwin2021chemformer,mahmood2021masked}, never on both.
This semantic gap persists across architectural flavors (e.g.,~GANs~\citep{mendez2020novo}, RL~\citep{born2021datadriven}, VAEs~\citep{gomez2018automatic}, GNNs~\citep{maziarz2022learning,mahmood2021masked}, flow~\citep{shi2020graphaf,jain2022biological} and diffusion models~\citep{xu2022geodiff}).
To our knowledge, all existing approaches either tune task-specific heads~\citep{lu2022unified} or limit the communication between both modules to a reward/loss and thus fail to "entangle" constrained structure generation with property prediction.
This critically violates the intuitive expectation that a property-driven generative model should, in the first place, excel at recognizing this property.

In this paper, we aim to close this gap by reformulating regression as a sequence modeling task. 
We propose the Regression Transformer (RT), a novel multitask model that can be trained on combinations of numerical and textual tokens (see~\autoref{fig:graphicalabstract}).
This circumvents the canonical way of addressing regression in Transformers, i.e., tuning a designated regression head~\citep{devlin2019bert}.
Despite solely relying on tokenization of numbers and cross-entropy loss, the RT can successfully solve regression tasks.
Notably, the same model can conditionally generate text sequences given continuous properties.
This is achieved simply by moving the \texttt{[MASK]} location and does not require finetuning specific heads; thus constituting a true multitask model.
To equip the RT with an inductive bias for handling floating-point properties, numbers are first tokenized into a sequence of tokens preserving the decimal order.
We then devise numerical encodings to inform the model about the semantic proximity of these tokens.
To allow for concurrent optimization of regression and conditional generation, we derive a PLM-inspired, alternating training scheme that includes a novel self-consistency loss for improved text generation based on continuous primers.
In the remainder of this paper, we describe the capabilities of the RT 
on a diverse set of predictive and generative tasks in chemical and protein language modeling.
We commence with small-molecule modeling, validate the RT on a synthetic dataset of drug-likeness~\citep{bickerton2012quantifying} and then test it on three property prediction datasets from the MoleculeNet benchmark~\citep{wu2018moleculenet}.
The property predictions results are compared with previous approaches relying on a regression loss and demonstrate that regression can be cast as conditional sequence generation task without losing accuracy.
These experiments rely on SELFIES~\citep{krenn2020self}, a chemical language devised for generative tasks that, as we show, has comparable predictive power to SMILES.
Although we aim to concurrently excel at predicting properties and generating sequences conditioned on properties, we start training with the PLM objective~\citep{yang2019xlnet} which does not explicitly model those tasks.
We then refine this objective and devise a training scheme that alternates between optimizing property prediction and text generation.
For the latter, we derive a novel self-consistency loss that exploits the dichotomy of the RT by querying itself with the generated candidate sequence.
To assess performance in conditional sequence generation, we systematically vary the continuous properties of interest and investigate the model's ability to adapt a seed sequence according to the primed property value.
We show applications on property-driven local chemical space exploration by decorating scaffolds with a continuum of properties and evaluate the novel molecules using the RT itself as well as an independent property predictor~\citep{rong2020self}.
The RT is then challenged against specialized molecular generative models on a property-driven molecular generation benchmark~\citep{jin2018junction}, where it significantly outperforms prior art. \\
Next, the RT is investigated on protein sequence modeling where it matches the performance of conventional Transformers on two regression datasets from TAPE~\citep{rao2019evaluating}.
In experiments on chemical reactions, we notice that the RT constitutes a generalization of forward reaction and retrosynthesis models.
We then demonstrate on two reaction datasets that the RT can not only predict reaction yields with similar accuracy to conventional Transformers~\citep{schwaller2021prediction}, but that it can also substitute specific precursors and thus generate novel reactions with higher predicted yield than a seed reaction. 


%

\section{Results}

\begin{figure}
    \caption{Results on chemical language modeling.}
    \label{fig:clm_results}
\begin{minipage}{0.55\textwidth}
\subfloat[
\textbf{Performance after PLM training.}
RMSE ($\downarrow$) and PCC (Pearson correlaction coefficient) refer to predicting QED, perplexity ($\downarrow$) to the PLM objective (\autoref{eq:plm_partial}) and Spearman $\rho$ ($\uparrow$) and $0$-Var ($\downarrow$) to the conditional generation task.
All values are means across multiple models.
Full table with standard deviations in appendix~\autoref{tab:plmqed_std}.
NE refers to the use of numerical encodings.
]{
\begin{footnotesize}
\begin{tabular}{cc|c|cc|cc}
\multicolumn{2}{c|}{\textit{Configuration}} & Per- & \multicolumn{2}{c|}{\textit{Regression task}} & \multicolumn{2}{c}{\textit{Generation task}}   \\
Data & NE & plexity ($\downarrow$) & RMSE ($\downarrow$) & PCC ($\uparrow$) & $0$-Var ($\downarrow$) & $\rho$ ($\uparrow$)    \\
\cline{1-7}
SMILES & -- & \ $\mathbf{1.55}$ & $0.0549$ & $\mathbf{0.972}$ & $1.6\%$ & $0.096$ \\
SELFIES & -- & $1.61$ & $0.0591$ & $0.968$ & $0.9\%$ & $0.427$\\
SELFIES & \cmark & $1.59$ & $\mathbf{0.0547}$ & $0.971$ & $\mathbf{0.3}\%$ & $\mathbf{0.467}$ \\
\cline{1-7}
\end{tabular}
\end{footnotesize} 
\label{tab:plmqed}
}
\end{minipage}
\qquad
\begin{minipage}{0.4\textwidth}
\subfloat[
\textbf{Performance evaluation on refined objectives.}
Legend like in Figure~\ref{tab:plmqed}.
NE means numerical encodings and $\alpha$ refers to the self-consistency loss function in~\autoref{eq:sc_objective}.
All models here used SELFIES, but the SMILES models as well as the full table with standard deviations can be seen in appendix~\autoref{tab:refined_qed_two_col}.
]{
\begin{footnotesize}
\begin{tabular}{cc|cc|cc}
\multicolumn{2}{c|}{\textit{Config}} & \multicolumn{2}{c|}{\textit{Regression task}} & \multicolumn{2}{c}{\textit{Generation task}}   \\
NE & $\alpha$  & RMSE & PCC & $0$-Var & Spearman $\rho$
\\
\cline{1-6}

\xmark & 0 & $\textbf{0.0341}$ & $\textbf{0.988}$  & $0.2\%$ & $0.47$ \\
\xmark & 1  & $0.0483$ & $0.978$  & $0.3\%$  & $0.49$ \\

\cmark & 0 & $0.0498$ & $0.982$ & $0.3\%$ & $0.47$\\
\cmark & 1 & $0.0367$  & $0.987$ & $\mathbf{0.2}\%$ &  $\mathbf{0.52}$ \\ 

\cline{1-6}
\label{tab:refined_qed}
\end{tabular}
\end{footnotesize} 
}
\end{minipage}


\begin{minipage}{0.3\textwidth}
\captionsetup{font=footnotesize}
\subfloat[
Performance comparison in predicting QED.
MAE stands for mean absolute error. 
The RT with alternating objectives used $\alpha=0$ in~\autoref{eq:sc_objective}.
]{
\begin{footnotesize}
\begin{tabular}{cc}
Model & MAE ($\downarrow$)   \\
\cline{1-2}
\textit{k}-NN (baseline) & $0.054$ \\
SMILES-BERT~\citep{kim2021merged} & 0.020 \\
\textbf{RT - PLM obj.} & 0.035 \\
\textbf{RT - Alternating obj.} &  \textbf{0.017} \\
\cline{1-2}
\end{tabular}
\label{tab:qedcomp}
\end{footnotesize} 
}
\end{minipage}
\qquad
\begin{minipage}{0.6\textwidth}
\begin{small}
\vspace{-1mm}
\subfloat[
\textbf{Learned embeddings of numerical tokens.}
\textit{Left:} For an exemplary dimension, embeddings for $20$ tokens, corresponding to $10$ digits and $2$ decimal places are shown. 
\textit{Right:} Embeddings for $20$ exemplary dimensions across $10$.
The stars indicate the significance level of the Pearson correlation.
The analysis is based on a SELFIES model without static NEs (PLM objective).
]{
\includegraphics[width=1.02\linewidth]{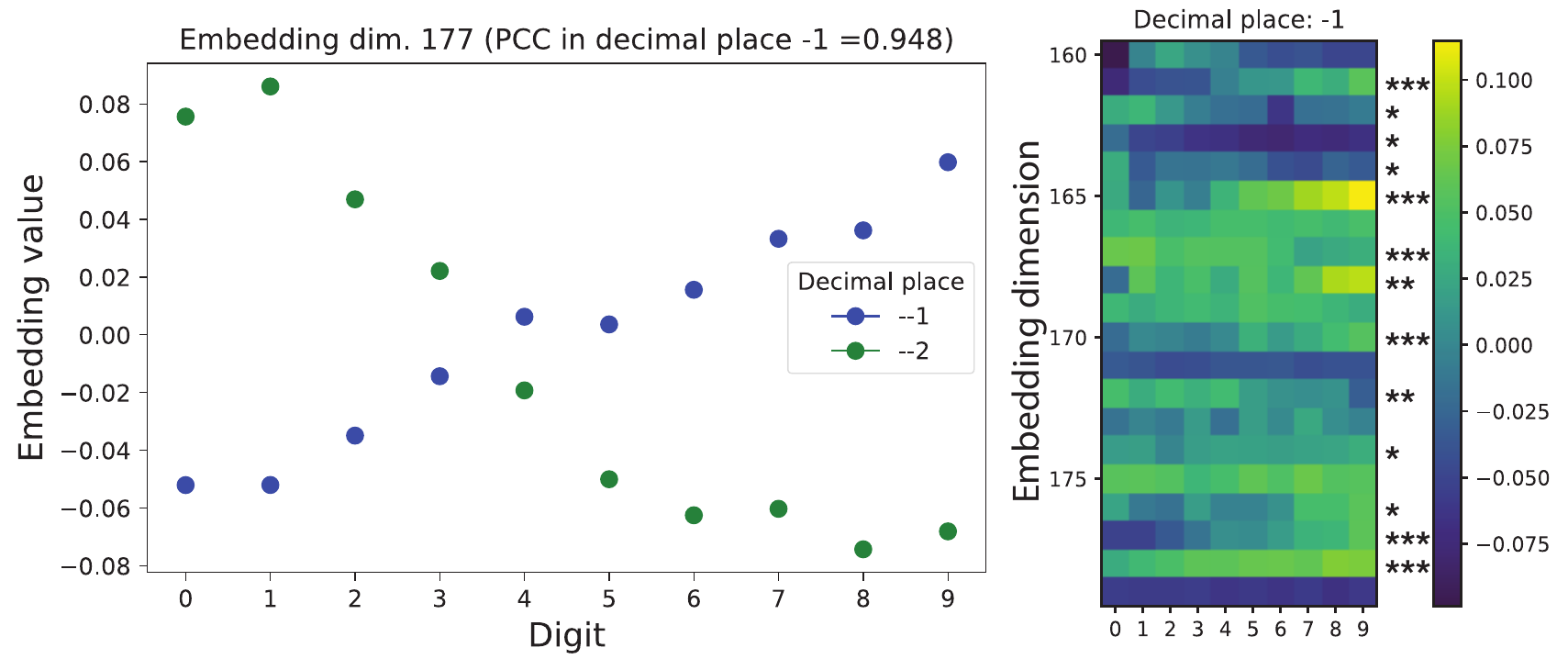}
\label{fig:learnednumericals}
}
\end{small}
\end{minipage}

\begin{minipage}{0.55\textwidth}
\hspace{-5mm}
\subfloat[
\textbf{RMSE ($\downarrow$) in predicting MoleculeNet dataset properties.}
Performance on three different datasets across predictive models.
By $\mathcal{L}_{Reg}$ we denote whether a given model used a loss (or objective function) that relied on regression.
All models used repeated random splits. 
NE means numerical encodings and $\alpha$ refers to the loss function in~\autoref{eq:sc_objective}.
]
{
\begin{footnotesize}
\begin{tabular}{cc|cccc}
Model & $\mathcal{L}_{Reg}$ & ESOL & FreeSolv & Lipo. \\
\cline{1-5}
RF~\citep{wu2018moleculenet} & \cmark & $1.16\pm_{0.15}$ & $2.12\pm_{0.68}$ & $0.78\pm_{0.02}$ \\
XGBoost~\citep{wu2018moleculenet} & \cmark & $1.05\pm_{0.10}$ & $1.76\pm_{0.21}$ & $0.84\pm_{0.03}$ \\
MPNN~\citep{wu2018moleculenet} & \cmark & $0.55\pm_{0.02}$ & $1.20\pm_{0.02}$ & $0.76\pm_{0.03}$ \\
\hdashline
SMILES-BERT~\citep{kim2021merged}  & \cmark & $\textbf{0.47}\pm_{0.05}$ & $\textbf{0.81}\pm_{0.09}$ & -- \\
Mol-BERT~\citep{fabian2020molecular} & \cmark &$0.53\pm_{0.04}$ & $0.95\pm_{0.33}$ & $0.56\pm_{0.03}$ \\
XLNet (ours) & \cmark &$0.69\pm_{0.01}$ & $1.03\pm_{0.25}$ & $\mathbf{0.74}\pm_{0.02}$ \\
\hdashline
\textbf{RT} ($\alpha=0$, NE: \xmark)  & \xmark & $0.76\pm_{0.05}$ & $1.19\pm_{0.29}$ & $0.76\pm_{0.03}$ \\
\textbf{RT} ($\alpha=1$, NE: \xmark)  & \xmark  & $0.75\pm_{0.04}$ & $1.32\pm_{0.39}$ & $0.76\pm_{0.03}$ \\
\textbf{RT} ($\alpha=0$, NE: \cmark)  & \xmark & $0.71\pm_{0.04}$ & $1.40\pm_{0.47}$ & $0.74\pm_{0.05}$ \\
\textbf{RT} ($\alpha=1$, NE: \cmark)  & \xmark & $0.73\pm_{0.04}$ & $1.34\pm_{0.29}$ & $0.74\pm_{0.03}$
\label{tab:moleculenet_prop}
\end{tabular}
\end{footnotesize} 
}
\end{minipage}
\begin{minipage}{0.5\textwidth}
\subfloat[
\textbf{Conditional generation for MoleculeNet datasets.}
Average performances across three splits for training with alternating objectives. 
$\rho$ refers to Spearman rank correlation and was evaluated with Grover~\citep{rong2020self}.
Same legend like Figure\autoref{tab:moleculenet_prop}.
Full table with standard deviations and self-evaluation with RT are in appendix~\autoref{tab:moleculenetcgappendix}.
]{
\begin{footnotesize}
\begin{tabular}{ccc|cccccc}
\multirow{2}{*}{Model} & & & \multicolumn{2}{c}{\underline{ESOL}} & \multicolumn{2}{c}{\underline{FreeSolv}} & \multicolumn{2}{c}{\underline{Lipophilicity}} \\
& NE & $\alpha$ & $0$-Var & $\rho$ & $0$-Var & $\rho$ & $0$-Var & $\rho$ \\
\cline{1-9}
\textbf{RT} & \xmark & 0 & $\textbf{4.4}\%$ & $0.44$ & $7.9\%$ & $0.53$ & $3.6\%$ & $0.29$ \\
\textbf{RT }& \xmark & 1 & $5.9\%$ & $0.46$ & $7.5\%$ & $0.56$ & $\textbf{2.7\%}$ & $\textbf{0.35}$ \\
\textbf{RT} & \cmark & 0 & $6.1\%$ & $0.46$ & $8.9\%$ & $\textbf{0.57}$ & $4.2\%$ & $0.29$ \\
\textbf{RT} & \cmark & 1 & $6.1\%$ & $\textbf{0.47}$ & $\textbf{6.5\%}$ & $\textbf{0.57}$ & $\textbf{2.7\%}$ & $0.34$ 
\label{tab:moleculenetcg}
\end{tabular}
\end{footnotesize} 
}
\end{minipage}

\end{figure}

\subsection{Chemical language modeling}
\subsubsection{Initial validations -- learning drug-likeness}
%
%
To test the feasibility of concurrent property prediction and conditional generation, we start with optimizing the vanilla permutation language objective (\autoref{eq:plm}) on a synthetic QED dataset (see~\autoref{fig:summary} for an illustration of how the mixed alphanumeric sequences are tokenized and embedded).
Since this objective masks tokens randomly in the sequence, evaluating such models on property prediction (i.e., masking only numerical tokens;~cf.~\autoref{fig:graphicalabstract}b \textit{top}) does not closely mimic their training dynamics.
Despite this (and the unconventional formulation of a regression task as sequence modeling), all models generated sequences of numerical tokens that allowed decoding floats, and even achieved a RMSE $<0.06$ (cf. Figure\autoref{tab:plmqed}).
%
%

%
%
Instead, for the generative task, the same models were queried $10$ times for every validation molecule with property primers\footnote{Throughout this manuscript by "primers" we mean that we replace the true property of a sequence with a desired property value.} equidistantly spaced in $[0,1]$ and $40\%$ of masked textual tokens.
The high rank correlation $\rho$ (between primers and QED of unique, generated molecules) values show that the model learned successfully to complete the corrupted scaffolds to produce full molecules with a desired QED.
Here, the SELFIES models exceeded the SMILES models by far, because SMILES, unlike SELFIES, can be syntactically invalid.
Due to the comparable results for property prediction (cf. Figure\autoref{tab:plmqed}), the remaining experiments focus exclusively on SELFIES.
Notably, the novelty score (i.e., percentage of conditionally generated molecules not present in training data) was $>99\%$ for all models. 
This demonstrates that the RT can generate novel chemical matter that adheres to a continuous property of interest.
Moreover, the numerical encodings (NE) slightly improved performance in all tasks.
Further ablation studies on different types of NEs and related work on encoding numbers with Transformer are reported in appendix~\ref{sec:ablationNE}.

\begin{figure*}[!htb]
\centering
\includegraphics[width=1.0\linewidth]{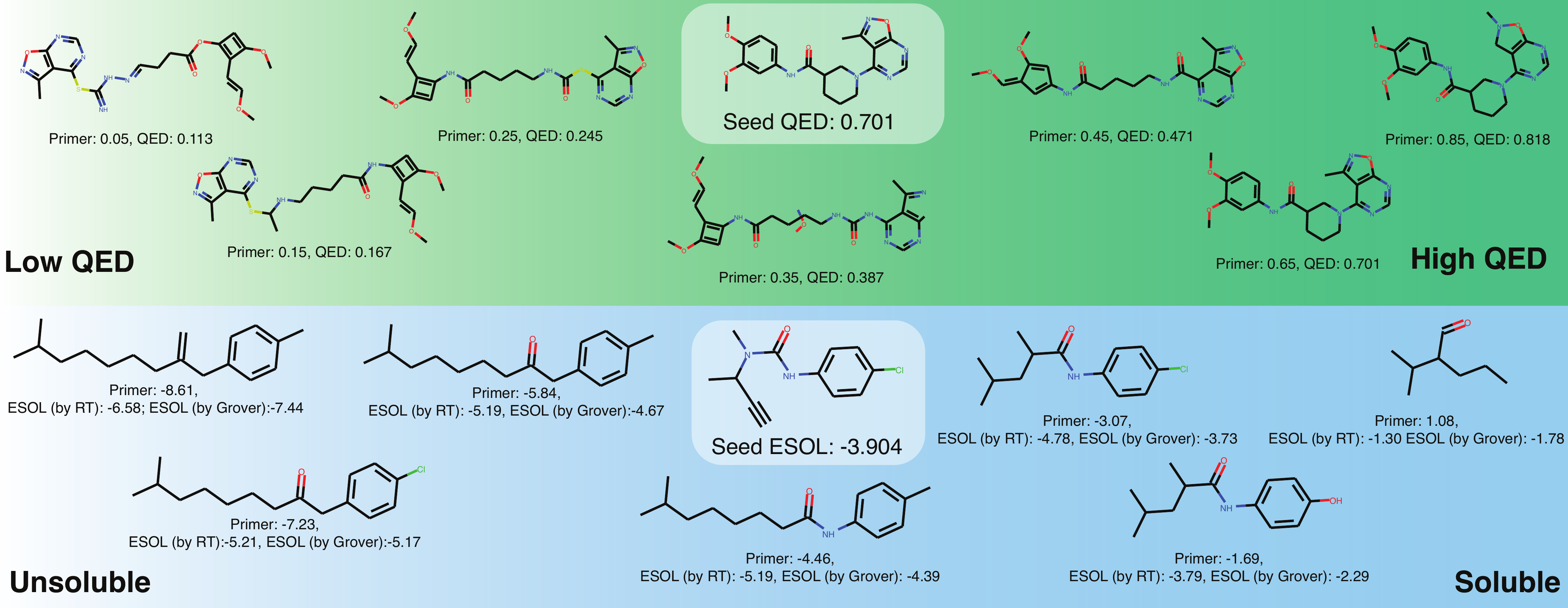}
\caption{
\begin{small}
\textbf{Property-driven, local optimization of molecular design with the Regression Transformer (RT).}
For each row, the seed molecule is shown in the middle alongside its true property. 
Based on $10$ property primers, $10$ molecules were decoded but duplicates were discarded.
Samples generated with the self-consistency model.
\textit{Top:} QED dataset.
\textit{Bottom:} ESOL dataset of aquatic solubility.
The solubility of the novel molecules was predicted by the RT itself and is externally validated by Grover~\citep{rong2020self}. 
\end{small}
}
\label{fig:mols}
\end{figure*}

%
%
Next, based on our proposed training scheme with alternating objectives, the models were refined:
For every model in Figure\autoref{tab:plmqed}, two models were trained, \textit{without} ($\alpha=0$) and \textit{with} ($\alpha=1$) the self-consistency term in the text loss (cf.~\autoref{eq:sc_objective}), respectively.
As shown in Figure\autoref{tab:refined_qed}, the performance in regression as well as conditional generation improved significantly, demonstrating the effectiveness of the refined objectives.
Moreover, all configurations of the Regression Transformer (RT) outperformed a baseline $k$-NN-regressor on Tanimoto similarity 
and our best configuration even surpassed the SMILES-BERT model~\citep{wang2019smiles} which achieved a MAE of 0.02 after pretraining on $\sim$9M SMILES with a regular regression loss~(see Figure\autoref{tab:qedcomp}).
%
%
%
%
The self-consistency term further improved the model's ability to generate tailored ensembles of molecules and led to consistently higher correlation scores.
This is exemplarily visualized in~\autoref{fig:mols} (\textit{top}) where a single seed molecule is decorated according to the property primers to cover the full range of QED scores.
Generally, the better performance of the self-consistency models ($\alpha=1$) in the generative tasks comes at the cost of slightly inferior regression performance (cf. Table\autoref{tab:refined_qed}).
Presumably, this is because the model weights in charge of the regression are confounded with the gradients from the self-evaluation (cf. Equation~\ref{eq:sc_objective}).
The novelty scores for the molecules generated in this setting were even slightly higher than for the PLM training ($>99.3\%$ for all models). 
A particularly challenging application for property-driven, local exploration of the chemical space is scaffold hopping; for an example on this see appendix~\ref{sec:scaff}.
For ablation studies on SMILES language and other types of numerical encodings, see appendix~\ref{sec:ablationNE}.

\paragraph{Learning embeddings of numbers.}
We sought to understand why the ablation studies on the numerical encodings (NE) on the QED dataset (Table~\ref{tab:plmqed} and~\ref{tab:refined_qed}) reveal only mild but not enormous superiority of models with NEs.
%
%
%
Interestingly, in the absence of static NEs, the model learns the natural ordering of digits from the data (cf. Figure~\ref{fig:learnednumericals}).
A large number of embedding dimensions ($47\%$ and $36\%$ for the decimal places $-1$ and $-2$ respectively) directly and significantly encoded the ordering of digits (i.e., $p<0.05$ and $|PCC|>0.62$ between the $10$ embedding values and a strictly monotonic vector). 
For example, in Figure~\ref{fig:learnednumericals} (\textit{left}) the digit value is monotonically related to its embedding value.
Notably, this ordering trend was much less present in the models using NEs ($\sim16\%$).
For reference, with random weights, $5\%$ would be expected.\\
In general, attention weights in Transformers can capture complex semantics such as protein folding structure~\citep{vig2021bertology} or atom-mapping in chemical reactions~\citep{schwaller2021extraction}.
For a qualitative comparison of the RT's attention across the predictive and generative task, see appendix~\ref{sec:attention}.


\subsection{Regression benchmark (MoleculeNet) }
After the successful initial experiments, we evaluated the RT on three regression benchmarks from MoleculeNet~\citep{wu2018moleculenet}.
The regression performance on ESOL, FreeSolv and Lipophilicity is shown in Table~\ref{tab:moleculenet_prop} and compared to prior work.
The strongest baseline model from MoleculeNet, XGBoost, is outperformed by all our models on all tasks.
Even the MPNN~\citep{gilmer2017neural}, a message-passing GNN, is slightly surpassed on FreeSolv and Lipophicility by some of our models.
However, all our models are outperformed by BERT-based approaches~\citep{wang2019smiles,kim2021merged}.
Notably, these models leveraged large-scale self-supervised pretraining before finetuning a regression head.
Since these results might not be directly comparable to the RT with its XLNet backbone, we also finetuned a XLNet model with a conventional regression head.
Notably, despite the absence of a regression los, the RT is on par (\textit{Lipophilicty}) or only mildly inferior (i.e., within standard deviation range;~\textit{ESOL}, \textit{FreeSolv}) to XLNet.

But in stark contrast to all those approaches, only the RT can also be used to conditionally \textit{generate} molecules similar to the training samples~(cf.~Figure\autoref{tab:moleculenetcg}).
%
%
%
Since the properties of the generated molecules are intractable to evaluate \textit{in-silico}, we could predict them, handily, using the RT.
However, as this might be a biased estimator, 
we evaluated them using Grover~\citep{rong2020self}, a self-supervised Graph Transformer.
Hence, the Spearman reported in Figure\autoref{tab:moleculenetcg} is based on Grover's predictions.
Overall, the generative results underline the benefit of the self-consistency loss ($\alpha=1$) and demonstrate that the RT can adapt unseen seed molecules even according to complex molecular properties like water solubility.
For a qualitative evaluation, we depict the generations for one exemplary seed molecule of the solubility dataset in~\autoref{fig:mols} (\textit{bottom}).
Last, corroborative for our work was the high correlation of our property predictions (RT) with Grover's for molecules generated by the ESOL, FreeSolv and Lipo models ($0.86$, $0.84$ and $0.75$ respectively).
Thus, the Spearman $\rho$ scores obtained with RT predictions are consistent to Grover (cf.~\autoref{tab:moleculenetcgappendix}).

\subsection{Conditional molecular generation benchmark}
To assess whether the RT is a powerful conditional generative model,
we benchmarked it on a property-driven molecular generation task, namely pLogP constrained optimization~\citep{jin2018junction}.
Given a seed molecule and a similarity constraint to the seed molecule ($\delta$, given in Tanimoto similarity), the goal is to generate molecules with higher pLogP values.
The results in~\autoref{tab:propopt} demonstrate that, for both similarity thresholds $\delta$, the RT obtained the best results.
Across both similarities, it outperforms a Junction-Tree-VAE~\citep{jin2018junction} and a GCPN by $614\%$ and $103\%$ in average improvement, respectively.
\begin{table}[!htb]
\caption{
\textbf{Constrained property optimization benchmark.}
GCPN stands for graph-convolutional policy network~\citep{you2018graph}.
}
\label{tab:propopt}
\centering
\begin{footnotesize}
\subfloat[Similarity threshold $\delta=0.4$]{
\begin{tabular}{c|ccc|c}
& \multicolumn{3}{c|}{\textit{Generation task}} & \textit{Regression}\\
Model & Improvem. & Similarity $\delta$  & Success  & PCC \\
\hline
JT-VAE~\citep{jin2018junction} & $0.84_{\pm_1.5}$ & $0.51_{\pm0.1}$ & $83.6\%$ & \textit{Unfeasible} \\
GCPN~\citep{you2018graph} & $2.49_{\pm1.3}$ & $0.47_{\pm0.1}$ & $\textbf{100}\%$ & \textit{Unfeasible}\\
\textbf{RT (Ours)}  & $\textbf{3.16}_{\pm1.5}$ & $\textbf{0.54}_{\pm0.1}$ & $97.1\%$ & $\textbf{0.92}_{\pm 0.0}$ \\
\hline
\end{tabular}
}
\end{footnotesize} 
\vspace{-7mm}
\label{tab:plogpfirst}
%
\centering
\begin{footnotesize}
\subfloat[Similarity threshold $\delta=0.6$]{
\begin{tabular}{c|ccc|c}
& \multicolumn{3}{c|}{\textit{Generation task}} & \textit{Regression}\\
Model & Improvem. & Similarity $\delta$  & Success  & PCC \\
\hline
JT-VAE~\citep{jin2018junction} &  $0.21_{\pm0.7}$ & $\textbf{0.69}_{\pm0.0}$ & $46.4\%$ & \textit{Unfeasible}\\
GCPN~\citep{you2018graph} & $0.79_{\pm0.6}$ & $0.68_{\pm0.1}$ & $\textbf{100}\%$ & \textit{Unfeasible}\\
\textbf{RT (Ours)}  & $\textbf{2.21}_{\pm1.3}$ & $\textbf{0.69}_{\pm0.1}$ & $81.8\%$ &  $\textbf{0.92}_{\pm 0.0}$  \\
\hline
\end{tabular}
}
\end{footnotesize} 
\label{tab:plogpsecond}
\end{table}
While the success rate of GCPN is higher than ours, we emphasize that both JT-VAE and GCPN applied gradient optimization schemes at \textit{inference time}. 
Instead, the RT does not only not require any optimization at this stage, but it was also never trained explicitly to produce molecules with high pLogP.
This finding demonstrates that the RT is able to compete with specialized conditional generative models in goal-directed molecular generation.
At the same time, the RT also predicted the pLogP value with a Pearson's correlation of $0.92$, a task that cannot be addressed with normal conditional generative models.
The results in~\autoref{tab:propopt} were obtained with the RT including a self-consistency loss, but for ablation studies on the RT and further results on $\delta=0.2$ and $\delta=0$, see appendix~\ref{sec:ablation_propopt}.
%
%
%
%
%
\subsection{Protein sequence language modeling}
\subsubsection{Synthetic pretraining: Potential-protein-interaction (Boman index)}
To assess the generality of the RT beyond chemical languages,
we benchmarked the RT in protein language modeling.
On the synthetic pretraining data, the RT obtained nearly perfect results in predicting Boman's index (Spearman $\rho>0.994$;~Table\autoref{tab:protregression})
\begin{table}[t]
\captionof{table}{
Results on protein language modeling.
}
\begin{minipage}{0.45\textwidth}
\vspace{2mm}
\subfloat[\textbf{Protein regression tasks.}
All values in Spearman's $\rho$ ($\uparrow$) on the test set.
TAPE datasets/performances taken from~\cite{rao2019evaluating}.
An ablation study on the three loss functions (Equations~\ref{eq:plm},~\ref{eq:cg_objective} and~\ref{eq:sc_objective}) confirmed the superiority of the self-consistency objective~(see appendix~\ref{sec:ablation_boman} and~\autoref{tab:bomanablation}).
]{
\begin{tabular}{ccccc}
Model & Source & Boman & Fluoresc. & Stability \\
\cline{1-5}
$k$-NN & Baseline & $0.93$ & $0.59$ & $0.21$ \\
One-Hot & TAPE & -- & $0.14$ & $0.19$ \\
LSTM & TAPE & -- & $0.67$ & $0.69$ \\
Transformer & TAPE & -- & $0.68$ & $\textbf{0.73}$ \\
UniRep & \citep{alley2019unified} & -- & $0.67$ & $\textbf{0.73}$ \\
\textbf{RT} & \textbf{Ours} &  $\textbf{0.99}_{\pm0.01}$ & $\textbf{0.72}_{\pm0.04}$ & $0.71_{\pm 0.02}$  \\
\cline{1-5}
\label{tab:protregression}
\end{tabular}
}
\end{minipage}
\hfill
\begin{minipage}{0.45\textwidth}
\vspace{-5mm}
\hspace{-5mm}
\subfloat[\textbf{Protein generation tasks.}
Standard deviations measured across three runs. 
Ablation studies on the different loss functions are in appendix~\ref{sec:ablation_boman}.
]{

\begin{tabular}{c|cc|cc}
\multirow{2}{*}{Model} & \multicolumn{2}{|c|}{Boman dataset} & \multicolumn{2}{|c}{Stability dataset} \\
 & $0$-Var ($\downarrow$) & Spearm. $\rho$ & $0$-Var ($\downarrow$) & Spearm. $\rho$ \\
\cline{1-5}
All TAPE & \multicolumn{2}{c|}{\multirow{2}{*}{\textit{Task unfeasible}}}  &  \multicolumn{2}{c}{\multirow{2}{*}{\textit{Task unfeasible}}} \\
UniRep &  &   \\ 
\textbf{RT} &  $0.2\%_{\pm 0.0}$ & $0.84_{\pm0.00}$ &  $19\%_{\pm 4.5}$ & $0.44_{\pm0.01}$ 
\label{tab:protgeneration}
\end{tabular}
}
\end{minipage}
\end{table}
and outperformed a baseline $k$-NN using Levenshtein distance~\citep{levenshtein1966binary}.
But the RT also successfully generated peptides with a desired Boman index, given a partially corrupted amino-acid sequence (cf.~Spearman $\rho$ of $0.84$, see Table\autoref{tab:protgeneration}).
Also, a higher fraction of masked tokens lead to better results in protein generation tasks~(cf.~appendix~\autoref{fig:boman_ablation}).

\subsubsection{TAPE datasets (protein fluorescence \& protein stability)}
Next, the RT performed competitively on two realistic protein regression datasets from TAPE~(cf.~Table\autoref{tab:protregression}).
This is remarkable given that the TAPE models were pretrained large-scale on unlabelled protein sequences and finetuned with a regression loss.
For example, the RT outperforms all reported methods in Spearman correlation on the Fluorescence task; which has a distribution with two modes, for bright and dark proteins respectively.
Inspecting the predictions in more depth showed that the RT excels at recognizing the mode of a protein but struggles with intra-mode precision (see appendix~\ref{sec:fluor}). 
Overall, the competitive predictive performance of the RT demonstrates that the benefits of self-supervised pretraining can extend to numerically labelled datasets.
This yields, \textit{en passant}, a conditional generative model for property-driven local exploration of the protein sequence space.
Evidence on this can be found in Table\autoref{tab:protgeneration}: 
Whereas all TAPE models as well as the UniRep method are incapable of addressing this generation task, the RT was able to modify the test proteins such that their (predicted) stability correlated strongly with the primed property ($\rho=0.44$). 

\FloatBarrier
\subsection{Modeling chemical reactions}
Language models advanced reaction chemistry significantly~\citep{schwaller2018found,schwaller2021extraction} and also showed superior performance on yield prediction~\citep{schwaller2021prediction}, yet models incorporating yield into (partial) reaction generation are lacking entirely.

\begin{figure}[tbp]
\caption{Chemical reaction modeling. }
\centering
    \vspace{1mm}
    \hspace{-5mm}
    
\begin{minipage}{0.45\textwidth}
    \subfloat[Reaction yield prediction performance for ten 70/30 splits, measured in coefficient of determination (\textit{R}\textsuperscript{2}).]{
\begin{tabular}{c c  c }
\multirow{2}{*}{Model} & Buchwald- & Suzuki-\\
  & Hartwig  & coupling \\ \hline
One-Hot~\citep{sandfort2020structure} & $0.89$ & --\\ 
DFT~\citep{ahneman2018predicting}    & $0.92$      & -- \\
MFF~\citep{sandfort2020structure} & $0.927_{\pm0.01}$ & --\\ 
Yield-BERT~\citep{schwaller2021prediction} & $\textbf{0.951}_{\pm0.01}$ & $0.79_{\pm0.02}$\\ 
Yield-BERT finetuned & $\textbf{0.951}_{\pm0.01}$ & $\textbf{0.81}_{\pm0.01}$\\ 
RT (\textbf{ours}) & $0.939_{\pm0.01}$ & $\textbf{0.81}_{\pm0.02}$\\ \hline
\vspace{-3mm}
\label{tab:yield_predict}
\end{tabular}
}
\end{minipage}
    \hspace{-5mm}
    \hfill
    \begin{minipage}{0.56\textwidth}
    \raggedleft
\subfloat[Generating precursors for Buchwald-Hartwig aminations~\citep{ahneman2018predicting}.
Each reaction in the dataset also contained 4-Methylaniline and the same Palladium-catalyst, thus they are excluded from the analysis. 
For legend, please see Table~\ref{tab:yield_suz}.
]{
\vspace{-5mm}
\begin{tabular}{ c|  c  c| c  c}
\multirow{3}{*}{} & \multicolumn{2}{c|}{\textit{Reconstruction}} & \multicolumn{2}{c}{\textit{Decoration}} \\
 Precursor & Top-3  & Similarity & Success & Mean \\
  & accuracy  & $\delta$ & rate & improv. \\ \hline 
Halide
& $98.23\%_{\pm{0.5}}$ & $0.991_{\pm{0.00}}$ & $42.3\%_{\pm{2.4}}$ & $6.1_{\pm{1.3}}$ \\
Ligand 
& $50.38\%_{\pm{1.6}}$ & $0.677_{\pm{0.01}}$ & $74.4\%_{\pm{4.2}}$ & $14.4_{\pm{1.7}}$ \\
Base 
& $100\%_{\pm{0.0}}$ & $1.000_{\pm{0.00}}$ & $82.2\%_{\pm{2.3}}$ & $8.1_{\pm{0.6}}$ \\
Additive & $1.36\%_{\pm{0.5}}$ & $0.158_{\pm{0.02}}$ &$71.2\%_{\pm{1.8}}$ & $11.7_{\pm{1.3}}$  
\label{tab:yield_buch}
\end{tabular}
}
\end{minipage}

    \begin{minipage}{.8\textwidth}
    \centering
    \subfloat[Generating precursors for Suzuki couplings~\citep{perera2018platform}.
    Each reaction in the dataset also contained the same Palladium-catalyst which is thus excluded from this analysis. 
Full precursors were generated ($p_{mask}=1$). 
For \textit{reconstruction}, we show the percentage of cases where the exact right precursor was among the top-3 predicted sequences and the Tanimoto similarity of the most similar of those molecules. For \textit{decoration}, we show the percentage of cases where the top-5 predicted reactions contained a reactions with higher (predicted) yield than the seed reaction (succes rate), alongside the associated average yield improvement.
]{
\begin{tabular}{ c|  c  c|  c  c}
\multirow{3}{*}{Precursor} & \multicolumn{2}{|c|}{\textit{Reconstruction}} & \multicolumn{2}{c}{\textit{Decoration}} \\
 & Top-3  & Similarity & Success & Mean  \\
  & accuracy  & $\delta$ & rate & improv. \\ \hline
Electroph. & $44.2\%_{\pm{17.6}}$ & $0.732_{\pm{0.02}}$ & $63.5\%_{\pm{7.1}}$ & $12.5_{\pm{3.4}}$ \\
Nucleoph.   & $100.0\%_{\pm0.0}$  & $1.000_{\pm 0.00}$ & $54.0\%_{\pm{6.2}}$ & $5.4_{\pm{0.8}}$ \\
Ligand & $67.4\%_{\pm{20.0}}$ & $0.689_{\pm{0.15}}$ & $56.7\%_{\pm{3.5}}$ &  $5.5_{\pm{0.6}}$ \\
Base & $90.5\%_{\pm{1.2}}$ & $0.811_{\pm{0.01}}$ & $47.8\%_{\pm{2.7}}$ & $4.6_{\pm{0.3}}$ \\
Solvent & $56.4\%_{\pm{1.1}}$ & $0.661_{\pm{0.01}}$ & $57.8\%_{\pm{1.8}}$ & $7.5_{\pm{0.3}}$ 
\label{tab:yield_suz}
\end{tabular}
}
  \end{minipage}
  \begin{minipage}[b]{0.8\columnwidth}
    \subfloat[Together with a BH amination from the validation dataset (\textit{top}), we show two RT-generated reactions with adaptations of the base and halide respectively, both with higher (predicted) yield by the RT.
    The RXN confidence stems from the forward model by~\citet{schwaller2019molecular} which confirmed that the reaction would result in the shown product in all cases.
    For improvements of additive and ligand of the same reaction, please see~\autoref{fig:yield_example_full}.
    ]{
    \label{fig:yield_example_main}
    \hspace{-5mm}
    \includegraphics[width=1.0\textwidth,left]{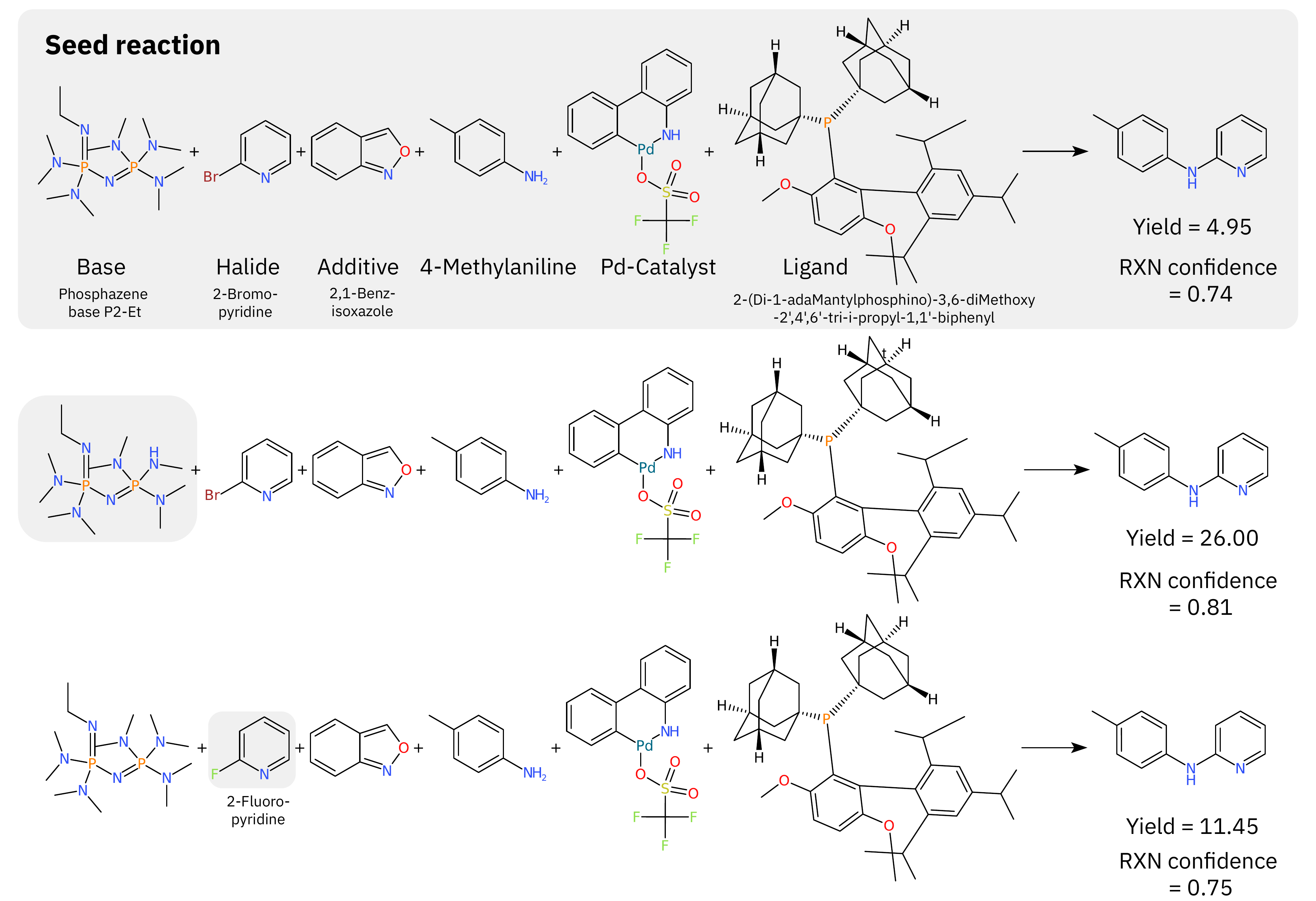}
    }
    \end{minipage}

\end{figure}

We therefore optimized the RT for concurrent yield prediction and precursor generation on two reaction-yield datasets: Buchwald-Hartig aminations~\citep{ahneman2018predicting} and Suzuki-Miyaura cross-couplings~\citep{perera2018platform}. 
On yield prediction, the RT (trained on SELFIES) outperforms fingerprint-based or quantum-mechanics methods, and matches (Suzuki dataset) or almost matches (Buchwald dataset) the performance of language models like Yield-BERT, trained with regression loss on SMILES~(cf.~Table\autoref{tab:yield_predict}).
%
%
%
%
%
%




%
The same model learned to reconstruct missing precursors in Buchwald-Hartwig animations which can be useful to infer missing solvents or reagents in automatically extracted reactions~(cf.~Table\autoref{tab:yield_buch}).
This is partly achieved with great accuracy (e.g.,~$98.2\%$ for aryl-halides).
Interestingly, inferring additives proved challenging, possibly because they are the dominant precursor type for the reaction yield~\citep{ahneman2018predicting}.
However, upon masking the additive only partially (rather than completely), the reconstruction performance increases significantly~(ablation study with $p_{mask}\in [0.25, 0.5, 1]$ in~\autoref{tab:buchwald_additive}).
On the Suzuki-couplings, the reconstruction results are more balanced among the five precursor types; the average Tanimoto similarity to the true precursor was $>0.65$ in all cases (cf. Table\autoref{tab:yield_suz}).
Moreover, across both datasets we observed mild benefits in reconstruction performance when providing the true yield rather than masking it (cf.~\autoref{tab:yield_buch_extended}/~\autoref{tab:yield_suz_extended}).
In addition to yield prediction and precursor reconstruction, the RT can also \textit{decorate} existing reactions by adapting specific precursors toward a higher yield (cf.~Table\autoref{tab:yield_buch}/\autoref{tab:yield_suz})).
Consistently among both datasets and all precursor types, 40-80\% of the top-5 predicted sequences contained reactions with entirely novel precursors and higher predicted yield.

Figure\autoref{fig:yield_example_main} visualizes exemplary adaptations of base and arly-halide of a BH amination with very low yield ($<5\%$).
Notably, for this unseen reaction, the RT found novel adaptations of each of the four precursor types that resulted in an increase of predicted yield by $11$-$85\%$ (see~\autoref{fig:yield_example_full} for full details).
With the forward reaction prediction model in \texttt{IBM RXN}~\citep{schwaller2019molecular} we confirmed that all reactions indeed result in the desired product. 
Notably, the confidence from the forward model rank-correlated almost perfectly with the yield predicted by the RT ($\rho=0.90, p<0.05$).

\section{Discussion}
Here, we have presented the Regression Transformer (RT), demonstrated that regression can be casted as conditional sequence learning task and introduced a flexible multitask-language-model with wide application in scientific discovery.
Our main contribution is a "swiss army knife" transformer that bridges previously considered disjoint tasks (property prediction and conditional generation), excels at both tasks and could thus pave the road toward foundation models in material design.

Regarding molecular property prediction, we find that the RT learns continuous properties even from small datasets, surpasses conventional regression models on several benchmarks
and sometimes competes with Transformers trained on regression loss.
Remarkably, this is achieved without providing ratio-scale information about the property, potentially even challenging the necessity of using regression rather than classification objectives.

The experiments on conditional text generation underline the versatility of the RT: 
Across a wide range of tasks, we conditionally generated novel sequences (molecules, proteins, reactions) that seemingly adhere to primed, continuous properties.
We foresee this to be useful for property-driven, sub-structure constrained molecular or protein design.
Our experiments on constrained molecular generation benchmark further demonstrate that the RT can surpass specialized conditional generative models.

Moreover, we emphasize that even though all experiments reported herein examined singular properties, the RT naturally scales to multiproperty prediction (see "Software" section on how to access pretrained multiproperty models).

Future work could, for example, intensify the work on reaction modeling (the RT effectively  generalizes forward reaction and retrosynthesis models) or improve the ability of the RT to perform fine-grained regression (for an interesting failure mode see appendix~\ref{sec:failure}).
Finally, our work resonates with the recent trend towards multitask Transformers~\citep{sanh2021multitask,lu2021pretrained,brown2020language} and we envision it as a mean to accelerate the development of foundation models for scientific discovery applications.



\section*{Software and Data}
\subsection*{Reproduction}
The codebase to facilitate reproduction of all experiments is publicly available at:
\url{https://github.com/IBM/regression-transformer}.\\
\subsection*{Data}
The data for the MoleculeNet experiments can be obtained from: \url{https://moleculenet.org/datasets-1} \\
The data for the molecular optimization experiments can be obtained from:
\url{https://github.com/wengong-jin/icml18-jtnn/tree/master/data/zinc} \\
The data for the protein language modeling experiments can be obtained from: \url{https://github.com/songlab-cal/tape} \\
The data for the reaction yield experiments can be obtained from: 
\url{https://github.com/rxn4chemistry/rxn_yields/tree/master/data} \\

\subsection*{Usage of trained models}
The RT is implemented in the Generative Toolkit for Scientific Discovery (GT4SD~\citep{manica2022gt4sd}) which provides ready-to-use pipelines for inference and training/finetuning on custom data.
Via GT4SD, versions of the RT trained on the QED and ESOL datasets (small molecules) and the stability dataset (proteins) are available.
Moreover, GT4SD also distributes two additional versions of the RT trained on multi-property prediction tasks.
A notebook with a short demo can be found under:~\url{https://github.com/GT4SD/gt4sd-core/blob/main/notebooks/regression-transformer-demo.ipynb}.
The datasets used for benchmarking are available from the respectively referenced papers.

\newpage
\section{Methods}
\subsection{XLNet backbone}
The Regression Transformer (RT) is built upon an XLNet backbone~\citep{yang2019xlnet} to retain the benefits of auto-regressive modeling in combination with a bidirectional context.
At its core, XLNet is an auto-regressive language model but due to its novel training objective, it, in expectation, obtains full bidirectional attention. 
This bidirectionality is critical because the RT is required to fill multiple tokens at arbitrary positions in a sequence while attending the full remaining sequence\footnote{e.g. SMILES/SELFIES are non-local sequences such that masking functional groups usually implies masking disconnected tokens.}.
Moreover, the independence assumption in bidirectional but non-autoregressive models (like BERT) becomes increasingly disruptive as more masked tokens are filled, making XLNet the best choice.
This limits BERT's applicability for generative tasks in biochemistry like scaffold decoration where large portions of a molecule might be masked and generation of individual atoms can critically alter the molecule's functional properties.
In general, it is important to notice that the proposed framework can be applied to all transformer flavors, but it certainly benefits from an autoregressive generation with full sequence attention even for discontiguous mask locations, like XLNet or MPNet~\citep{song2020mpnet}.
%
%
\subsection{Tokenization}
This section describes the processing of alphanumeric sequences, i.e., strings consisting of a mixture of numerical and textual symbols~(for a visualization of the tokenization see~\autoref{fig:summary}, \textit{top}).
Unlike previous approaches that modelled 8-bit integers (i.e., pixels~\citep{van2016pixel}) with a classifier, we strive to represent real numbers with arbitrary floating point precision.
Since representing every number as a single token is suboptiomal due to a lack of generalization to new numbers and sparsity of the provided tokens, we formulated regression as sequential categorical task.
In turn, this necessitates a scheme for converting text representing numbers into a sequences of tokens.
First, the following regular expression splits a string denoting a numerical:
\begin{small}
\begin{equation}
    \verb/\s*\s*?(\+|-)?(\d+)(\.)?(\d+)?\s*/
    \label{eq:regexp}
\end{equation}
\end{small}
Each of the resulting matches containing a number is converted to a token $t_{v,p}$ where $v\in \mathbb{N} \cap [0..9]$ is the value/digit and $p\in\mathbb{Z}$ is the decimal place (e.g., $12.3$ is split into [\texttt{1\_1}, \texttt{2\_0}, \texttt{.}, \texttt{3\_-1}]).
We call these \textit{numerical tokens}.
This representation has the advantage that it allows easy decoding of the digit sequence but also distinguishes their decimal order by adhering to classic positional notation.
Negative numbers are preceded with a special token.
Regarding alphabetic tokens, we represent molecules as SELFIES~\citep{krenn2020self} strings and tokenized them with their internal tokenizer. 
In one ablation study, we instead use SMILES~\citep{weininger1988smiles} and tokenize with the regular expression from~\citet{schwaller2018found}.
Protein sequences are tokenized per amino acid.

\subsection{Numerical encodings (NE)}
Due to the inherent structure of numbers, learning the embeddings of numerical tokens in a purely data-driven way might be ineffective.
Moreover, since the RT is trained with cross-entropy loss, no notion of similarity between numerical tokens is conveyed.
As a remedy, we propose numerical encodings (NE), a simple inductive bias about the semantic proximity of numerical tokens, similar to positional encodings~\citep{vaswani2017attention}.
In practice, we sum the NEs with regular word embeddings and relative positional encodings from XLNet~(see Appendix~\autoref{fig:summary} for a workflow).
%
Our proposed numerical encodings are zero vectors for all but numerical tokens of the dictionary.
We follow positional notation as above.
Given a token $t_{v,p}$ (with digit value $v$ and decimal place $p$), the numerical encoding at embedding dimension $j$ is defined as:
\begin{equation}
NE_{Float}(v, p, j) = (-1)^{j} \cdot \frac{v\cdot10^p}{j{+}1}
\label{eq:float-encodings}
\end{equation}
Thus, the amplitude of the NE scales with the numerical value of the token.
The NEs are perfectly correlated among embedding dimensions but alternate between positive and negative values for even and odd dimensions and vanish for higher dimensions (see example in Figure~\ref{fig:floatencodings}\textit{a}).
Critically, the pairwise distances of the numerical encodings are symmetric and decay monotonically with the float value (see Figure~\ref{fig:floatencodings}\textit{b}).
Note that we also experimented with integer-based numerical encodings~(cf.~Supplementary Material~\ref{sec:numerical_encoding} for additional experiments).

\subsection{Training objectives}
The input $\mathbf{x}$ for a RT is defined by a concatenation of $k$ property tokens $[\mathbf{x}^p]_k$ and $l$ textual tokens $[\mathbf{x}^t]_l$, such that:~$\mathbf{x}=[\mathbf{x^p},\mathbf{x^t}]_T=[x_1^p,...,x_k^p, x_1^t,...,x_l^t]_T$. The full sequence length is $T{=}k{+}l$ and $\mathbf{x}^p$ and $\mathbf{x}^t$ are property and textual tokens respectively.

%
%
\paragraph{Permutation language modeling (PLM) objective.}
The idea of PLM~\citep{yang2019xlnet} is to fill masked tokens auto-regressively by sampling a factorization order $\mathbf{z}$ for a sequence $\mathbf{x}$ at runtime. Decomposing the likelihood $p_\theta(\mathbf{x})$ according to the facorization order yields, in expectation, a bidirectional auto-regressive model.
Let $\mathbf{z}\in \mathcal{Z}_T$ denote one of the $T!$ permutations of our sequence $\mathbf{x}$. 
If $z_i$ and $\mathbf{z}_{<i}$ are the $i$-th and first $i-1$ elements of $\mathbf{z}$, the PLM objective is:
\begin{equation}
    \max_\theta \mathbb{E}_{\mathbf{z}\sim\mathcal{Z_T}}\left[\sum_{i=1}^T \log p_\theta(x_{z_i}|\mathbf{x}_{\mathbf{z}_{<i}})\right]
    \label{eq:plm}
\end{equation}
In practice, partial prediction is performed, i.e., only the last $c$ tokens of the factorization order $\mathbf{z}$ are predicted.
Following XLNet, $\mathbf{z}$ is split into a (masked) target subsequence $\mathbf{z}_{>c}$ and an unmasked input sequence $\mathbf{z}_{\leq c}$ s.t. the objective becomes
\begin{equation}
\begin{aligned}
    \mathcal{J}_{PLM} = \max_\theta \mathbb{E}_{\mathbf{z}\sim\mathcal{Z_T}}\left[\log p_\theta(\mathbf{x}_{\mathbf{z}>c}|\mathbf{x}_{\mathbf{z}_{\leq c}})\right] \\ = \mathbb{E}_{\mathbf{z}\sim\mathcal{Z_T}} \left[\sum_{i=c{+}1}^T \log p_\theta(x_{z_i}|\mathbf{x}_{\mathbf{z}_{<i}}) \right]
\end{aligned}
    \label{eq:plm_partial}
\end{equation}
where $c$ is a hyperparamter, usually sampled per batch such that the fraction of masked tokens is roughly $1/c$.
We notice that (\ref{eq:plm_partial}) does not make any specific choices on $\mathbf{x}^p$ and $\mathbf{x}^t$. 
It thus constitutes our baseline objective. 
While (\ref{eq:plm_partial}) is a generic objective, it is computationally exhaustive to optimize due to the permutations. 
Moreover it is not ideal for our needs because it does not distinguish between textual and property tokens. 
Instead, we are aiming to develop a single model that can either predict numerical tokens (when given text sequences) or text tokens (when given a combination of numerical and text tokens).
To that end, we propose to train on two alternating objectives, one designed for property prediction and one for text generation. 

\paragraph{Property prediction objective.}
Instead of randomizing which tokens are masked, this objective exclusively masks all the property tokens.
Specifically, we constrain the factorization order $\mathbf{z}$ by setting the first $l$ elements to $\mathbf{x^t}$ and fixing $c=l$. 
This guarantees that only property tokens are masked.
Let $\mathcal{Z}_T^p$ denote the set of possible permutations.
Under this constraint, then the objective becomes
\begin{equation}
\begin{aligned}
    \mathcal{J}_{P} = \max_\theta \mathbb{E}_{\mathbf{z}\sim\mathcal{Z}_T^p}\left[\log p_\theta(\mathbf{x}^p|\mathbf{x}^t)\right] \\ = \mathbb{E}_{\mathbf{z}\sim\mathcal{Z}_T^p} \left[\sum_{i=c{+}1}^T \log p_\theta(x_{z_i}^p|\mathbf{x}^t_{\mathbf{z}_{\leq c}},\mathbf{x}^p_{\mathbf{z}_{>c<i}}) \right]
\end{aligned}
    \label{eq:prop_objective}
\end{equation}
where $\mathbf{x}^p_{\mathbf{z}_{>c<i}}$ denotes the $c$-th to the $i{-}1$-th element of the factorization order $\mathbf{z}$.
We emphasize that this "tailored" property objective $\mathcal{J}_p$ is still optimized with a cross-entropy loss in practice. 
Note that this loss cannot convey any notion on the qualitative proximity of the prediction to the labels because the level of measurement of tokens in a language model are on a nominal level.
Thus, predicting a sequence of numerical tokens corresponding to a property score of $0.91$ for a sample with a true property of $0.11$ will not generally result in a higher loss than predicting $0.21$.
Instead, a traditional regression loss operates on a ratio scale.

\paragraph{Conditional text generation objective.}
This objective facilitates the generation of textual tokens given a property primer and textual tokens.
We constrain the factorization order $\mathbf{z}$ by setting the first $k$ elements to $\mathbf{x}^p$ to and sampling the cutoff $c$, s.t. $c>=k$.
This ensures that masking only occurs on textual tokens.
With this constraint, we denote the set of permutations by $\mathcal{Z}_T^t$ and the objective becomes
\begin{equation}
\begin{aligned}
    \mathcal{J}_{G} = \max_\theta \mathbb{E}_{\mathbf{z}\sim\mathcal{Z}_T^t}\left[\log p_\theta(\mathbf{x}^t_{\mathbf{z}_{>c}}|\mathbf{x}^p_{\mathbf{z}_{\leq k}},\mathbf{x}^t_{\mathbf{z}_{>k<c}})\right] \\ = \mathbb{E}_{\mathbf{z}\sim\mathcal{Z}_T^t} \left[\sum_{i=c{+}1}^T \log p_\theta(x_{z_i}^t|\mathbf{x}^p_{\mathbf{z}_{\leq k}},\mathbf{x}^t_{\mathbf{z}_{>k<i}}) \right]
\end{aligned}
    \label{eq:cg_objective}
\end{equation}
Intuitively, this objective applies regular PLM while sparing the numerical tokens. 
It then aims to reconstruct the full text sequence (i.e., molecule) given the uncorrupted property tokens and partially corrupted textual tokens.
\paragraph{Self-consistency (SC) objective.}
Standalone, the above conditional text generation objective (\ref{eq:cg_objective}) does not reward if the generated sequences adhere to the primed property.
This is critical because in chemical as well as natural languages, changes in single tokens (i.e., atoms, amino acids or (sub)words) can drastically change the property (meaning) of a sequence (sentence).
As a remedy, we extended the text generation objective $\mathcal{J}_{G}$ by a self-consistency term 
that exploits the dichotomy of the Regression Transformer. 
The full objective is given by:
\begin{equation}
    \mathcal{J}_{SC} = \mathcal{J}_{G}(\mathbf{x}) + \alpha \cdot \mathcal{J}_{P}(\mathbf{\hat{x}})
    \label{eq:sc_objective}
\end{equation}
where the second addend is the self-consistency term, weighted by a factor $\alpha$.
Intuitively, it is given by the difference between the property of the sample and the predicted property of the generated sample $\mathbf{\hat{x}}$.
Here, $\mathbf{\hat{x}}$ is obtained by greedy decoding of the masked tokens and combining it with the non-corrupted tokens of $\mathbf{x}$.
To be precise, $\mathbf{\hat{x}}=[\mathbf{x}^p,\mathbf{\hat{x}}^t]$ where $\mathbf{\hat{x}}^t=[m_1 \bar{x}_1+(1{-}m_i)x_1, ..., m_l \bar{x}_l+(1{-}m_l)x_l]$. 
Here, $\mathbf{m}$ is an indicator vector whether masking occurred at a given position and $\mathbf{\bar{x}}=\arg \max \sum_{i=c+1}^T \log p_\theta(x_{z_i}^t|\mathbf{x}^p_{\mathbf{z}_{<k}}, \mathbf{x}^t_{\mathbf{z}_{>k<i}})$ is the result of greedy decoding.
In such a formulation, the RT acts as an oracle during its own optimization, resembling an additional layer of self-supervision.
While this scheme risks undesired side effects when the model performs poorly at property prediction, it introduces a notion of self-consistency and rewards the generation of molecules that are different from training samples as long as they adhere to the property.

%
%
\subsection{Evaluation \& performance metrics.}

\subsubsection{Regression.}
For the regression (or property prediction) task, we convert the sequence of predicted (numerical) tokens into a floating-point prediction (the model never failed to predict a token sequence not corresponding to a valid numerical). 
We then report the root-mean-squared error (\textbf{RMSE}), Pearson's correlation coefficient (\textbf{PCC}) or the coefficient of determination (\textbf{R\textsuperscript{2}}), dependent on the dataset and previous methods.

\subsubsection{Conditional sequence generation.}
Dependent on the application domain, different metrics are utilized. 
\vspace{-2mm}
\paragraph{Small molecule and protein modeling.}
We strive to assess the model's ability to decorate an arbitrary, possibly discontiguous fractional input sequence (e.g., a molecular scaffold) according to a property of interest.
Therefore, we randomly mask a fraction of tokens of the text sequence and then query the model with ten equidistant property primers spanning the full range of property values.
The metric is the average \textbf{Spearman's} $\bm{\rho}$  between the ten primers and the actual properties.
Spearman is favorable over Pearson because it is only rank-sensitive. 
Note that due to constraints induced by the fragmented sequence, covering the entire property spectrum is usually impossible such that e.g., RMSE is inappropriate for this task (e.g., priming a highly toxic scaffold with low toxicity cannot yield a non-toxic molecule).
As a sanity check, we also report $\mathbf{0}$\textbf{-Var}, i.e., the percentage of samples for which the generation was unaffected by the primer (the lower the better). \\
On the property optimization benchmark from~\citet{jin2018junction}, we report the same metrics as in their work.
The success rate in generating molecules with higher logP (while adhering to the similarity constraint $\delta$), the Tanimoto similarity $\delta$ to the seed molecule and the average improvement in logP.
\vspace{-2mm}
\paragraph{Chemical reaction modeling.}
For the reaction yield datasets, we challenge the model by two sequence generation tasks.
First, fully \textit{reconstructing} a precursor solely based on the remaining precursors and the reaction yield. 
The top-3 predicted sequences (decoded via beam search) are considered, s.t. \textbf{Top-3 accuracy} is reported.
Additionally we report the average \textbf{Tanimoto similarity} of the most similar of the top-3 molecules to the seed molecule (fingerprint: ECFP4).
Secondly, we measure the capability of \textit{decorating} existing reactions to obtain a (potentially) higher yield. 
To that end, the model is prompted with incomplete reactions consisting of an increased yield, an entirely masked precursor and complete remaining precursors.
We consider the top-3 predicted sequences (decoded via beam search) and report the fraction of samples where one of the reactions had a higher (predicted) yield (\textbf{success rate}). 
The second response metric is the \textbf{mean improvement} in (predicted) reaction yield (yield $y\in[0,100]$, the distributions are right-skewed).
Note that we exclude trivial solutions by removing all predicted precursors that exist in the training dataset.
\vspace{-2mm}

\subsection{Datasets}
\subsubsection{Chemical language modeling}
\paragraph{Synthetic QED dataset.}
Starting from $\sim1.6$M bioactive molecules from ChEMBL~\citep{mendez2019chembl}, we created a synthetic dataset by computing the QED~\citep{bickerton2012quantifying} score (q $\in[0,1]$) for all molecules with~\texttt{RDKit} and rounded to $3$ decimal places.
We used $\sim1.4$M molecules for training, $1$k for validation and $10$k for testing. 
\paragraph{MoleculeNet datasets.}
We focused on $3$ regression datasets from the MoleculeNet benchmark~\citep{wu2018moleculenet}: \textit{ESOL}, \textit{FreeSolv} and \textit{Lipophilicity}, where the task is to predict water solubility, hydration free energy and lipophilicity of a molecule, respectively.
For each dataset, we performed $3$ random splits (as recommended by~\cite{wu2018moleculenet}) with 15\% validation data.
Because the datasets are small ($<5000$ samples), we used SMILES augmentation~\citep{bjerrum2017smiles} to augment the dataset by a factor of $16$.
\paragraph{Property-optimization benchmark.}
This is a benchmark for property-driven, conditional molecular generation.
The goal is to adapt a seed molecule such that a property is maximized while adhering to a fixed similarity constraint.
We obtained the data from~\citet{jin2018junction} which ships with a fixed split of 215,381 training and 799 test molecules and their penalized LogP (pLogP) value~\citep{kusner2017grammar}.
pLogP is the octanol-water partition coefficient (logP) penalized by the synthetic accessibility score and the number of cycles with $>6$ atoms.
Hence, pLogP just like QED can be computed deterministically from the molecule. 
To maximize comparability we followed the candidate assembly process of~\citet{jin2018junction}, described in appendix~\ref{sec:propopt}.

\subsubsection{Protein sequence language modeling}
\paragraph{Synthetic Boman dataset.}
As a large-scale, labelled dataset we focused on the Boman index, a measure of potential protein interaction for peptides.
It is the average of the solubility values of the residues~\citep{boman2003antibacterial}.
We collected all 2,648,205 peptides with 15 to 45 AAs from UniProt~\citep{uniprot2021the}, computed their Boman index, and used 10k and 1k for testing and validation respectively.

\paragraph{TAPE datasets.}
We focused on two datasets from the TAPE benchmark~\citep{rao2019evaluating}:
\textit{Fluorescence}~\citep{sarkisyan2016local} and \textit{Stability}~\citep{rocklin2017global}. 
The goal is to predict, respectively, the fluorescence and intrinsic folding stability of a protein that is one to four mutations away from a training protein.
Both datasets ship with fixed splits. 
The fluorescence (stability) dataset has 21,446 (53,416) training, 5,362 (2,512) validation and 27,217 (12,851) test samples. 

\subsubsection{Chemical reaction datasets}
We investigated two high-throughput experimentation (HTE) yield datasets that examine specific reaction types: Buchwald-Hartig aminations~\citep{ahneman2018predicting} and Suzuki-Miyaura cross-coupling reactions~\citep{perera2018platform}.  
Both datasets were investigated in the same 10 random splits as examined in~\citet{schwaller2021prediction} with a 70/30\% train/validation ratio.

\paragraph{Buchwald-Hartwig.}
This dataset, produced by~\citet{ahneman2018predicting}, investigates HTE of Palladium-catalysed Buchwald-Hartwig C-N cross coupling reactions. 
The reaction space comprises $3955$ reactions, spanned by 15 unique aryl and heteroaryl halides, 4 Buchwald ligands, 3 bases and 22 isoxazole additives. A Palladium-catalyst and a Methylaniline are the fifth and sixth precursor respectively, however they are identical for all reactions.
Each reaction is associated to a yield $y \in [0,100]$ and the 10 random split were identical to the ones released by~\citet{sandfort2020structure} that are also used by all competing methods in Table\autoref{tab:yield_buch}. 
Yield is given in a range of $[0,100]$.

\paragraph{Suzuki cross-couplings.}
This dataset was provided by~\citet{perera2018platform} and investigates HTE of Suzuki-Miyaura reactions across $15$ pairs of electrophiles and nucleophiles, leading to different products respectively.
For each pair, a combination of 4 solvents, 12 ligands and 8 bases (reagents) was measured, resulting in a total of $5760$ reaction yields that we scale to the range $[0,100]$. 
The catalyst is identical for all reactions, some reactions omitted the ligand or the base while others contained electrophiles, nucleophiles, ligands, bases or solvents that were composed of different fragments (e.g., salts).

\paragraph{USPTO.}
Before training on the narrow yield datasets, we warmed up the model to learn generic reaction chemistry.
We used reactions from the US Patent Office (USPTO), the largest open-source dataset about chemical reactions~\citep{lowe2017chemical}. 
Since no yield information was available, the utilized numerical property was the total molecular weight of all precursors.
The dataset contained $n=2,830,616$ reactions and was obtained from~\citet{schwaller2021extraction}.

\section{Acknowledgements}
The authors would like to thank the entire IBM RXN for Chemistry team and especially Carlo Baldassari and Artem Leonov for useful discussions.

\newpage
\bibliography{references}
\bibliographystyle{unsrtnat}

\newpage

\renewcommand{\thepage}{A\arabic{page}}
\renewcommand{\thesection}{A\arabic{section}}
\renewcommand{\thetable}{A\arabic{table}}
\renewcommand{\thefigure}{A\arabic{figure}}
\setcounter{figure}{0}
\setcounter{table}{0}
\setcounter{page}{1}
\setcounter{section}{0}

\onecolumn
\clearpage
\pagebreak
\section*{\huge{Appendix}}

\section{Training and evaluation procedure}
All experiments build upon the XLNet~\citep{yang2019xlnet} backbone from the~\texttt{HuggingFace} library~\citep{wolf2019huggingface}. 
We expanded the XLNet backbone with our proposed tokenization scheme, an additional encoding layer for the numerical embeddings ($N_{dim}=16$) and the custom training objectives~(cf.~\autoref{fig:summary}).

{\begin{figure}[!htb]
\centering
\includegraphics[width=1\linewidth]{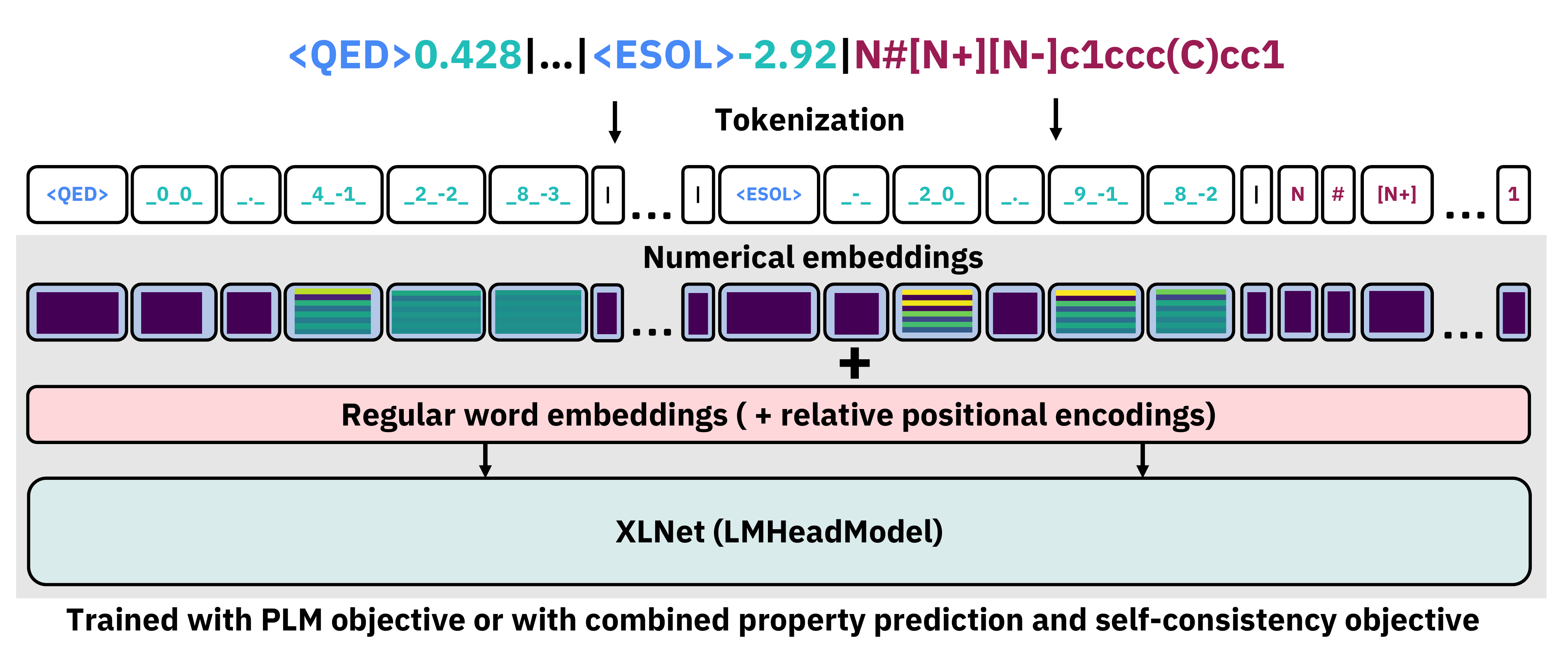}
\caption{
\begin{small}
\textbf{Workflow of the Regression Transformer (RT) model.}
Based on the XLNet backbone, the RT is a dichotomous model designed to handle combinations of text and numbers.
\textit{Top:} An input sequence consisting of a molecular string (red) and two property tags (blue), each associated to a floating value (green).
Numbers are tokenized into a sequence of tokens that preserve the decimal order of each character. 
The pipe (\texttt{|}) is a separator token distinguishing numerical and text tokens.
\textit{Middle}: We propose numerical encodings that inform the model about the semantic proximity of these tokens and naturally integrate with relative positional encodings and classical learned embeddings.
\textit{Bottom:} The RT is trained with an alternating training scheme, derived from the PLM objective~\citep{yang2019xlnet} and designed to concurrently optimize property prediction and conditional generation (\textit{bottom}).
The dots indicate that the RT naturally scales to multiple property tags.
\end{small}
}
\label{fig:summary}
\end{figure}}
%
%
Regarding architectural hyperparameters, we used $32$ hidden layers in the Transformer encoder, with a dimensionality of $256$ and $1024$ in the feed-forward layer and $16$ attention heads (20\% dropout).
Altogether, this model has $\sim 27$M trainable parameters (exact numbers vary dependent on vocabulary size).
During evaluation, greedy decoding was used for property prediction and beam search decoding for conditional sequence generation.
We used~\texttt{PyTorch} 1.3.1~\citep{paszke2019pytorch} and the XLNet backbone from \texttt{Transformers} 3.1.0~\citep{wolf2019huggingface}.
All models were trained on single GPUs~(\texttt{NVIDIA Tesla A100} or~\texttt{V100}).

In the following sections, we elaborate on the training procedures for each dataset.

\label{sec:training}
\subsection{Chemical language modeling}
\subsubsection{QED dataset.}
We started training the models with the vanilla permutation language modeling (PLM) objective (\autoref{eq:plm_partial}) on the QED dataset until validation perplexity saturated ($\sim4$ days, single-GPU).
Thereafter, the models were further refined on the same dataset by alternating every $50$ steps between objectives (\autoref{eq:prop_objective}) and (\autoref{eq:sc_objective}).
We perform ablation studies on the self-consistency loss, setting $\alpha$ in (\autoref{eq:sc_objective}) to $0$ and $1$ respectively.
The SELFIES/SMILES vocabulary had $509$ and $724$ tokens respectively.
During the latter, we gave the model more flexibility by setting $c=2.5$, s.t., $\sim40\%$ of the tokens were masked (maximum span: $7$ tokens).

\subsubsection{MoleculeNet dataset.}
For the MoleculeNet datasets, the models were warm-started using the QED initialization and trained only for $50$k steps (batch size $4$) with early stopping.
Since the QED pretraining utilized numerical values in $[0,1]$, we normalized the regression values of the MoleculeNet datasets to the same range and rounded them also to three decimal places.
For all objectives, unless otherwise constrained, we set the masking hyperparamter $c=5$ and restrict the span of consecutively masked tokens to a maximum of $5$ tokens.

\subsubsection{Property-optimization benchmark}
\label{sec:propopt}
For this task, the models were also warm-started using the QED initialization and trained for $50$k steps with early stopping on perplexity.
To assemble the candidates for the optimization of one seed molecule, we tried to follow the process of~\citet{jin2018junction} as closely as possible.
~\citet{jin2018junction} applied $80$ gradient steps, then decoded $80$ molecules and reported the molecule with the highest pLogP score that satisfies the similarity constraint $\delta$. 
Instead, we form a pool of molecules by prompting $80$ times with the same seed molecule but varying the fraction and the maximum span of masked tokens.
From the pool of decodings we report the molecule with the highest pLogP, just like~\citet{jin2018junction} and~\citet{you2018graph}.

\subsection{Protein sequence language modeling}
\subsubsection{Boman dataset}
To model protein sequences, we started with training on the Boman dataset.
We trained three groups of models, one for the vanilla PLM objective~(\autoref{eq:plm_partial}) and two for the alternating objectives.
We again alternated every $50$ steps between optimizing~(\autoref{eq:prop_objective}) and~(\autoref{eq:sc_objective}) and trained one set of models with and one set without the self-consistency loss, such that $\alpha=1$ and $\alpha=0$ respectively in~\autoref{eq:sc_objective}.
Models were trained until validation perplexity saturated ($\sim4$ days, single GPU).
The numerical values of the Boman index, originally in the range $[-3.1, 6.1]$ were normalized to $[0,1]$ and rounded to three decimal places.

\subsubsection{TAPE datasets}
Following the ablation study on the loss functions (see~\autoref{tab:bomanablation}) that revealed the best results for the self-consistency objective, we focused the finetuning exclusively on this configuration.
For both datasets, three models were warm-started using the Boman initialization and trained until validation performance saturated ($\sim100$k steps). 
The numerical values were again scaled to $[0,1]$.
On the Fluorescence data, a small value of Gaussian noise was added to some training samples due to an interesting failure mode (see~\ref{sec:failure}).
For the evaluation of the conditional generation task, the models were given more flexibility: $60\%$ of the tokens were masked (i.e., $c=1.7$ in~\autoref{eq:plm}) and the maximum span was $7$ AA residues.
We did not evaluate the RT on conditional generation for the Fluorescence dataset because of a massive pretraining-finetuning mismatch: While the Boman dataset used for pretraining consisted of $15$ to $45$ residues (mean/std: $36\pm7$), the fluorescence proteins were significantly larger ($246\pm0.2$ residues). 
Instead, the proteins in the stability dataset were similar in size to the pretraining data ($45\pm3$ residues).

\subsection{Reaction yield datasets}
\paragraph{Pretraining.}
Since the two reaction yield datasets only cover narrow regions of the chemical space (one template applied to many precursor combinations), we warmup the model on broader reaction chemistry extracted from patents (USPTO).
5000 reactions were held out for validation and the model was trained until validation performance on the two alternating objectives (\autoref{eq:prop_objective} and~\autoref{eq:sc_objective} with $\alpha=1$) saturated.
The masking hyperparameter $c$ was set to 2.5 and the model were trained for $\sim$ 2 days (single GPU).
The vocabulary for reaction SELFIES contained $861$ tokens.

\paragraph{Finetuning}
For both the Buchwald-Hartwig reactions~\citep{ahneman2018predicting} and the Suzuki-couplings~\citep{perera2018platform}, ten models were finetuned respectively on repeated random splits.
The training objectives again alternated every $50$ steps between property prediction (\autoref{eq:prop_objective}) and conditional generation (\autoref{eq:sc_objective} with $\alpha=1$) for a maximum of $50k$ steps ($\sim$~1 day).
Notably, during the conditional generation task we sampled one precursor per batch and then entirely but exclusively masked this precursor. 
Thus the objective for the model became to reconstruct a missing precursor from the remaining precursors and the reaction yield (or to produce an alternative precursor with a similar predicted yield).

\subsection{Baseline models.}
\subsubsection{$k$-Nearest-Neighbor ($k$-NN)}
For small molecule and protein modeling we reported results in property prediction with $k$-NN baseline model. 
For small molecules, the distance measure was (inverted) Tanimoto similarity~\citep{tanimoto1958elementary} of ECFP4 fingerprints~\cite{rogers2010extended}.
For the protein language models, the Levenshtein distance between the protein sequences was used~\citep{levenshtein1966binary}.
For the $k$-nn baseline models, $k$ was determined based on the best performance on the validation data. This led to $k=25$ for the drug-likeness/QED task, $k=21$ for the protein interaction (Boman index) task, $k=50$ for the fluorescence and $k=15$ for the stability task.

\subsubsection{XLNet with regression head}
For the molecular property prediction on the MoleculeNet datasets, we trained an XLNet~\citep{yang2019xlnet} model with a conventional regression loss. 
This maximizes comparability to the RT since it, unlike the other models in Table~\autoref{tab:moleculenet_prop}, also uses an XLNet backbone.
This model was initialized using the~\texttt{XLNet-base-cased} weights from~\texttt{HuggingFace} and subsequently the~\texttt{SequenceClassification} head was finetuned with an $L_2$ loss.
The model contained $\sim 93M$ parameters and was finetuned for $200$ epochs without any hyperparameter optimization. Early stopping was used to determine the best epoch.

\section{Numerical encodings}
\label{sec:numerical_encoding}
\subsection{Visualization of numerical encodings}
\begin{figure}[!htb]
\centering
\includegraphics[width=0.7\linewidth]{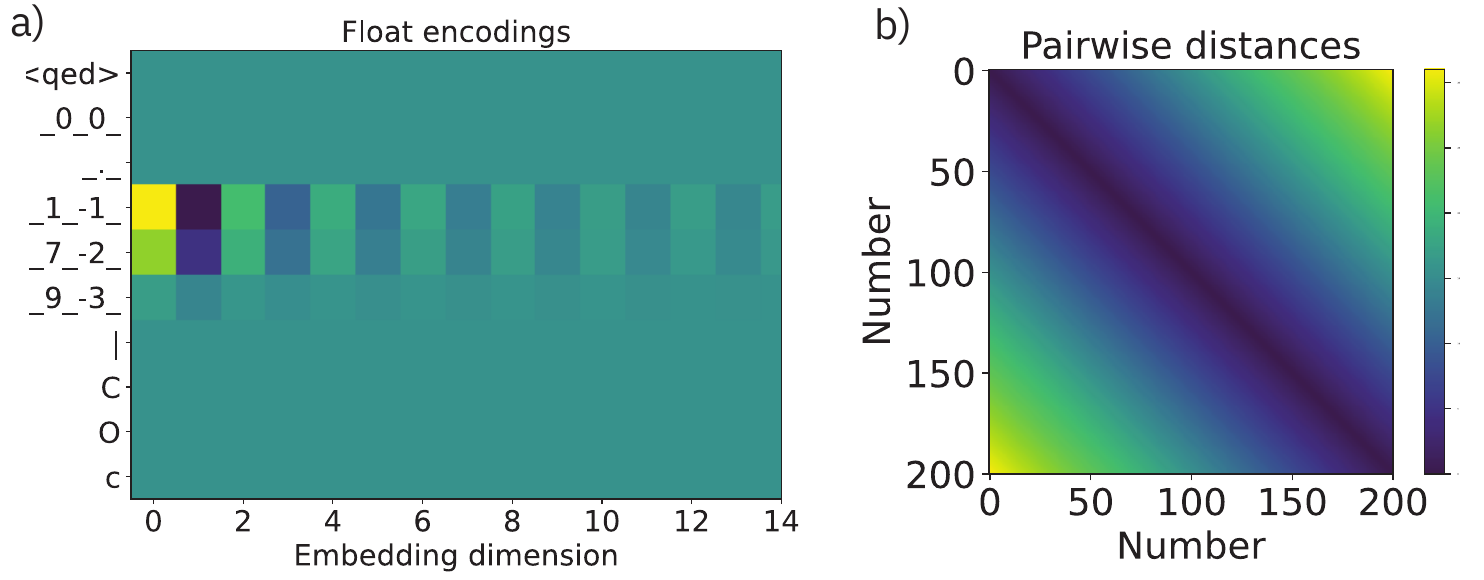}
\caption{
\begin{small}
\textbf{Float-based numerical encodings.}
\textit{a)} Numerical encodings for an molecule with a QED of $0.179$.
\textit{b)} Pairwise distances of numerical encodings for floats between $0$ and $100$ (the NEs of all tokens associated to a float are summed up).
\end{small}
}
\label{fig:floatencodings}
\end{figure}

\label{sec:ablationNE}
\subsubsection{Description of Integer encodings.}
\label{sec:intencoding}
As an alternative to the float-based numerical encodings (NE), we experimented with an encoding scheme relying solely on positive integers. 
Note that any regression problem 
can trivially be casted to a regression problem where all labels are positive integers.
Under this consideration, we need to define NEs only for positive integers\footnote{Strictly speaking only integers with a single, non-zero digit (i.e., covered by the base-10 exponentiation of the decimal system)}; similar to positional encodings.
We therefore propose to directly utilize the definition from~\citet{vaswani2017attention} as NEs:
\begin{equation}
\begin{split}
NE_{Int}(v,p, 2j) = \sin{\left[(v\cdot10^p)/10000^{2j/d_e}\right]} \\
NE_{Int}(v,p, 2j{+}1) = \cos{\left[(v\cdot10^p)/10000^{2j/d_e}\right]}
\label{eq:int-encodings}
\end{split}
\end{equation}
where $d_e$ is the embedding size.
The advantage of this integer-based encoding is that  every embedding dimension captures fluctuations of different frequencies; using trigonometric functions as continuous analogs to alternating bits. 
Practically, to use the Integer-NEs, the property values were casted to the range $[0,1000]$ and rounded.

\subsubsection{Float vs. Integer-based numerical encodings}
\label{sec:floatint}
\autoref{tab:plmqed_std} provides extended results of Table\autoref{tab:plmqed} in the main paper. 
\begin{table*}[!htb]
\centering
\captionsetup{font=footnotesize}
\captionof{table}{
\textbf{Performance evaluation of PLM training.}
FE refers our main float-encodings whereas Int refers to the Integer encodings described above.
}
\begin{footnotesize}
\begin{tabular}{ccc:cc:cc}
& & & \multicolumn{2}{:c:}{\textit{Regression task task}} & \multicolumn{2}{c:}{\textit{Generation task task}}\\
Data & NE & Perplexity ($\downarrow$) & RMSE ($\downarrow$) & PCC ($\uparrow$) & $0$-Var ($\downarrow$) & SCC ($\uparrow$)  \\
\cline{1-7}
SMILES & -- & \ $\mathbf{1.55}_{\pm0.02}$ & $0.0549_{\pm0.01}$ & $\mathbf{0.972}_{\pm0.01}$ & $1.6\%_{\pm0.2}$ & $0.096_{\pm0.02}$ \\
SELFIES & -- & $1.61_{\pm0.03}$ & $0.0591_{\pm0.00}$ & $0.968_{\pm0.00}$ & $0.9\%_{\pm0.2}$ & $0.427_{\pm0.01}$\\
SELFIES &FE & $1.59_{\pm0.03}$ & $\mathbf{0.0547}_{\pm0.01}$ & $0.971_{\pm0.00}$ & $\mathbf{0.3}\%_{\pm0.1}$ & $\mathbf{0.467}_{\pm0.01}$ \\
SELFIES & Int & $1.63_{\pm0.02}$ & $0.0564_{\pm0.00}$ & $0.968_{\pm0.00}$ &  $0.8\%_{\pm0.3}$ & $0.440_{\pm0.01}$\\
\cline{1-7}
\end{tabular}
\label{tab:plmqed_std}
\end{footnotesize} 
\end{table*}
It shows the standard deviations across several runs of the Regression Transformer.
In this setting, from the two types of proposed numerical encodings, the float-based encodings yielded slightly superior result to integer-based encodings.
Similarly,~\autoref{tab:refined_qed_two_col} shows extended results of~\autoref{tab:refined_qed} in the main paper, including standard deviations and the ablation study on integer vs. float encodings.
\begin{table*}[!htb]
\centering
\captionsetup{font=footnotesize}
\captionof{table}{
\textbf{Performance evaluation on alternating objectives.}
The decrease in perplexity compared to the vanilla PLM training is expected given the discrepancy between the refined, alternating objective and the PLM objective.
}
\begin{footnotesize}
\begin{tabular}{cccc:cc:cc}
& & & & \multicolumn{2}{:c:}{\textit{Regression task }} & \multicolumn{2}{c:}{\textit{Generation task}}\\
Data & NE & $\alpha$ & Perplexity & RMSE & Pearson's $r$ & $0$-Var & Spearman's $\rho$
\\
\cline{1-8}
SMILES & -- & $0$ & $2.15_{\pm0.1}$ & 0.0396 & 0.986 & $0.8\%_{\pm0.2}$ & $0.11_{\pm 0.02}$ \\
SMILES &-- & $1$ & $\mathbf{1.73}_{\pm0.1}$ & 0.0507 & 0.982 & $1.1\%_{\pm0.2}$ & $0.09_{\pm 0.02}$ \\
SELFIES & -- & $0$ & $2.57_{\pm0.1}$ & 0.0341 & 0.988  & $0.2\%_{\pm0.1}$ & $0.47_{\pm 0.02}$ \\
SELFIES &-- & $1$ & $2.41_{\pm0.1}$ & 0.0483 & 0.978  & $0.3\%_{\pm0.1}$  & $0.49_{\pm 0.01}$ \\
SELFIES & FE & $0$ & $2.10_{\pm0.1}$ & 0.0498 & 0.982 & $0.3\%_{\pm0.1}$ & $0.468_{\pm 0.03}$\\
SELFIES & FE & $1$ & $2.67_{\pm0.1}$ & 0.0367  & 0.987 & $\mathbf{0.2}\%_{\pm0.1}$ &  $\mathbf{0.52}_{\pm 0.02}$ \\ 

SELFIES & Int & $0$ & $2.63_{\pm0.1}$ & \textbf{0.0307} & \textbf{0.990} &  $0.7\%_{\pm0.2}$ & $0.41_{\pm 0.01}$ \\
SELFIES & Int & $1$ & $2.71_{\pm0.1}$ & 0.0412 & 0.986 &  $0.8\%_{\pm0.3}$ & $0.44_{\pm 0.01}$ \\ 
\cline{1-8}
\end{tabular}
\end{footnotesize} 
\label{tab:refined_qed_two_col}
\end{table*}
Here, integer-encodings (IE) are superior for regression but inferior for conditional generation.
Due to that and the non-applicability of IEs to floating numbers, we decided to not further explore them.

\paragraph{Summation vs. concatenation of numerical encodings.}
\label{sec:sumconcat}
We decided to follow the common approach of \textit{summing} additional encodings with the learned embeddings~\citep{vaswani2017attention,yang2019xlnet} but note that disentangling content and position embeddings can improve language models~\citep{he2021deberta}.
So, instead of summing the numerical encodings to the regular embeddings, we also experimented with concatenation (dimensionality of $32$ for the NEs.).
This produced slightly inferior but nearly identical results, see~\autoref{tab:qed_plm_ne}.
%
%
%
\begin{table}[!htb]
\centering
\captionsetup{font=footnotesize}
\captionof{table}{
\textbf{Ablation study on NEs.} 
Results on PLM training.
}
\begin{footnotesize}
\begin{tabular}{cccccccc}
NE & Type & RMSE ($\downarrow$) & PCC ($\uparrow$) \\
\cline{1-4}
-- & -- & $0.0591_{\pm0.00}$ & $0.968_{\pm0.00}$\\
Float & Concat. & $0.0581_{\pm0.00}$ & $0.966_{\pm0.01}$\\
Float & Sum & $\mathbf{0.0547}_{\pm0.01}$ & $\mathbf{0.971}_{\pm0.00}$\\
Int & Concat. & $0.0666_{\pm0.01}$ & $0.963_{\pm0.01}$\\
Int & Sum & $0.0564_{\pm0.00}$ & $0.968_{\pm0.00}$\\
\cline{1-4}
\end{tabular}
\end{footnotesize} 
\label{tab:qed_plm_ne}
\end{table}
We propose to use a summation for two reasons: 
First, it avoids additional hyperparameters and model weights.
Secondly, it probably yields approximately orthogonal subspaces of token embedding and numerical encodings (due to the high dimensionality).
This obviates the need to enforce orthogonality with a concatenation.
While we conjectured that using NEs improves the performance in both tasks (property prediction and conditional generation), we emphasize that providing this prior might not be necessary given enough data.
We hypothesize that refining our NEs might yield better results and in particular a faster convergence, but leave further refinement to future work, especially given the plethora of research about positional encodings~\citep{dai2019transformer,bai2021segatron,wang2020position}.

\newpage
\subsection{Conditional generation: External evaluation vs. self-evaluation}
Generally, it is intractable to evaluate the performance in most property-driven molecular generation tasks because the property of interest can only be measured in the wet lab.
In the main paper, we have reported the predicted ESOL, FreeSolv and Lipophilicity values respectively based on the GROVER approach~\citep{rong2020self}, a graph Transformer with large-scale self-supervised pretraining.
\autoref{tab:moleculenetcgappendix} shows that a self-evaluation with the Regression Transformer would have led to very similar results in all three conditional generation tasks.
This is reassuring because the RT is, at least in the self-consistency setting ($alpha=1$), a biased estimator since the model is used itself to optimize the conditional generation process.
Based on this finding, we refrained from seeking external validation for the conditional protein and reaction generation tasks.
\begin{table}[!ht]
\centering
\captionsetup{font=footnotesize}
\caption{
\textbf{Conditional generation for MoleculeNet datasets.}
Average performances across all splits for training with alternating objectives are given. 
"$\rho$ with RT" refers to the self-evaluation whereas "$\rho$ with Grover" refers to to predictions obtained with the model from~\citep{rong2020self}.
}
\label{tab:moleculenetcgappendix}
\begin{footnotesize}
\subfloat[\textbf{ESOL}]{
\begin{tabular}{ccccc}
Metric & $\alpha=0$, no FE  & $\mathbf{\alpha=1}$, no FE & $\mathbf{\alpha=0}$, with FE & $\mathbf{\alpha=1}$, with FE \\
\cline{1-5}
$0$-Variance ($\downarrow$) & $\mathbf{4.4}_{\pm0.8}$ & $5.9_{\pm1.3}$& $6.1_{\pm3.7}$ & $6.1_{\pm1.5}$ \\
Spearman $\rho$ based on RT predictions & $0.38_{\pm0.1}$ & $0.38_{\pm0.0}$& $0.41_{\pm0.1}$ & $\mathbf{0.44}_{\pm0.0}$ \\
Spearman $\rho$ with Grover predictions & $0.44_{\pm0.0}$ & $0.46_{\pm0.0}$& $0.46_{\pm0.1}$ & $\mathbf{0.47}_{\pm0.0}$ \\
\cline{1-5}
\end{tabular}
}
\end{footnotesize} 
\label{tab:cg_esol}
\end{table}
\begin{table}[!ht]
\ContinuedFloat
\centering
\begin{small}
\subfloat[\textbf{FreeSolv}]{
\begin{tabular}{ccccc}
Metric & $\alpha=0$, no FE  & $\mathbf{\alpha=1}$, no FE & $\mathbf{\alpha=0}$, with FE & $\mathbf{\alpha=1}$, with FE \\
\cline{1-5}
$0$-Variance ($\downarrow$)  & $7.9_{\pm2.4}$ & $7.5_{\pm3.6}$& $8.9_{\pm5.2}$ & $\mathbf{6.5}_{\pm2.6}$ \\
Spearman $\rho$ based on RT predictions & $0.51_{\pm0.0}$ & $\mathbf{0.52}_{\pm0.1}$& $\mathbf{0.52}_{\pm0.0}$ & $0.44_{\pm0.1}$ \\
Spearman $\rho$ based on Grover predictions & $0.53_{\pm0.0}$ & $0.56_{\pm0.0}$& $\mathbf{0.57}_{\pm0.0}$ & $\mathbf{0.57}_{\pm0.0}$ \\
\cline{1-5}
\end{tabular}}
\end{small} 
\label{tab:cg_freesolv}
\end{table}
\begin{table}[!ht]
\ContinuedFloat
\centering
\begin{small}
\subfloat[\textbf{Lipophilicity}]{
\begin{tabular}{ccccc}
Metric & $\alpha=0$, no FE  & $\mathbf{\alpha=1}$, no FE & $\mathbf{\alpha=0}$, with FE & $\mathbf{\alpha=1}$, with FE \\
\cline{1-5}
$0$-Variance ($\downarrow$) & $3.6_{\pm1.6}$ & $2.7_{\pm0.9}$& $4.2_{\pm1.3}$ & $\mathbf{2.7}_{\pm0.7}$ \\
Spearman $\rho$ based on RT predictions & $0.22_{\pm0.1}$ & $\mathbf{0.29}_{\pm0.0}$& $0.23_{\pm0.0}$ & $0.26_{\pm0.0}$ \\
Spearman $\rho$ based on Grover predictions & $0.29_{\pm0.1}$ & $\mathbf{0.35}_{\pm0.0}$& $0.29_{\pm0.0}$ & $0.34_{\pm0.0}$ \\
\cline{1-5}
\end{tabular}
}
\end{small} 
\end{table}

\newpage
\subsection{Conditional molecular generation (constrained property optimization benchmark)}
\label{sec:ablation_propopt}
On the constrained property optimization benchmark, we conducted ablation studies of the Regression Transformer for the use of  float-based numerical encodings (NE) as well as the self-consistency loss function. 
The main metric of this task is the mean improvement in pLogP compared to the seed molecule.
The results can be found in~\autoref{tab:ablation_propopt}.
The value of $\alpha$ refers to~\autoref{eq:sc_objective}: $\alpha=0$ means that no self-consistency loss was used and $\alpha=1$ implies that the self-consistency loss was used with a weight equal to the regular conditional text generation objective (cf.~\autoref{eq:cg_objective}).
The results of the ablation study indicate that the RT consistently outperformed the JT-VAE and GCPN in the main metric (improvement) by a wide margin.
\begin{table}[!ht]
\captionsetup{font=footnotesize}
\caption{
\textbf{Further results on constrained property optimization benchmark.}
JT-VAE is from~\citet{jin2018junction} and GCPN from~\citet{you2018graph}.
NE refers to the use of Numerical Encodings.
}
\label{tab:ablation_propopt}
\centering
\begin{footnotesize}
\subfloat[\textbf{No similarity threshold ($\delta=0.0$)}]{
\begin{tabular}{ccc:ccc:c}
\multicolumn{3}{c}{\textit{Training configuration}} & \multicolumn{3}{c:}{\textit{Generation task}} & \textit{Regression}\\
Model & Numerical Encoding & Self-consistency ($\alpha$) & \textbf{Improvement} & Similarity $\delta$  & Success rate  & Pearson's $r$ (PCC) \\
\hline
JT-VAE & -- & -- & $1.91_{\pm2.0}$ & $0.28_{\pm0.2}$ & $97.5\%$ & \textit{Unfeasible} \\
GCPN & -- & -- & $4.20_{\pm1.3}$ & $\textbf{0.32}_{\pm0.1}$ & $\textbf{100\%}$ & \textit{Unfeasible}\\
\textbf{RT} & \checkmark & $1$ &  $\textbf{8.67}_{\pm2.5}$ & $0.10_{\pm0.1}$ & $\textbf{100\%}$ & $0.92$ \\
\textbf{RT} & \cmark & $0$ &  $7.96_{\pm2.6}$ & $0.11_{\pm0.1}$ & $\textbf{100\%}$ & $0.90$ \\
\textbf{RT} & \xmark & $1$ &  $8.52_{\pm2.5}$ & $0.10_{\pm0.1}$ & $\textbf{100\%}$ & $0.91$ \\
\textbf{RT} & \xmark & $0$ &  $8.35_{\pm2.6}$ & $0.10_{\pm0.1}$ & $\textbf{100\%}$ & $\textbf{0.94}$ \\
\hline
\end{tabular}
}
\end{footnotesize} 
\label{tab:plogp_ablation0}
\end{table}
\begin{table}[!ht]
\centering
\begin{footnotesize}
\subfloat[\textbf{Similarity threshold $\delta=0.2$}]{
\begin{tabular}{ccc:ccc:c}
\multicolumn{3}{c}{\textit{Training configuration}} & \multicolumn{3}{c:}{\textit{Generation task}} & \textit{Regression}\\
Model & Numerical Encoding & Self-consistency ($\alpha$) & \textbf{Improvement} & Similarity $\delta$  & Success rate  & Pearson's $r$ (PCC) \\
\hline
JT-VAE & -- & -- & $1.68_{\pm1.9}$ & $0.33_{\pm0.1}$ & $97.1\%$ & \textit{Unfeasible} \\
GCPN & -- & -- & $4.12_{\pm1.2}$ & $0.34_{\pm0.1}$ & $\textbf{100}\%$ & \textit{Unfeasible}\\
\textbf{RT} & \checkmark & $1$ &  $\textbf{4.45}_{\pm1.7}$ & $0.35_{\pm0.1}$ & $99.6\%$ & $0.92$ \\
\textbf{RT} & \cmark & $0$ &  $4.12_{\pm1.7}$ & $\textbf{0.36}_{\pm0.1}$ & $99.6\%$ & $0.90$ \\
\textbf{RT} & \xmark & $1$ &  $4.34_{\pm1.6}$ & $0.35_{\pm0.1}$ & $99.9\%$ & $0.91$ \\
\textbf{RT} & \xmark & $0$ &  $4.40_{\pm1.7}$ & $0.35_{\pm0.1}$ & $99.7\%$ & $\textbf{0.94}$ \\
\hline
\end{tabular}
}
\end{footnotesize} 
\label{tab:plogp_ablation2}
\end{table}
\begin{table}[!ht]
\centering
\begin{footnotesize}
\subfloat[\textbf{Similarity threshold $\delta=0.4$}
]{
\begin{tabular}{ccc:ccc:c}
\multicolumn{3}{c}{\textit{Training configuration}} & \multicolumn{3}{c:}{\textit{Generation task}} & \textit{Regression}\\
Model & Numerical Encoding & Self-consistency ($\alpha$) & \textbf{Improvement} & Similarity $\delta$  & Success rate  & Pearson's $r$ (PCC) \\
\hline
JT-VAE & -- & -- & $0.84_{\pm1.5}$ & $0.51_{\pm0.1}$ & $83.6\%$ & \textit{Unfeasible} \\
GCPN & -- & -- & $2.49_{\pm1.3}$ & $0.47_{\pm0.1}$ & $\textbf{100}\%$ & \textit{Unfeasible}\\
\textbf{RT} & \checkmark & $1$ &  $\textbf{3.16}_{\pm1.5}$ & $0.54_{\pm0.1}$ & $97.1\%$ & $0.92$ \\
\textbf{RT} & \cmark & $0$ &  $2.87_{\pm1.5}$ & $\textbf{0.55}_{\pm0.1}$ & $95.5\%$ & $0.90$ \\
\textbf{RT} & \xmark & $1$ &  $3.09_{\pm1.5}$ & $0.54_{\pm0.1}$ & $97.2\%$ & $0.91$ \\
\textbf{RT} & \xmark & $0$ &  $3.04_{\pm1.5}$ & $0.54_{\pm0.1}$ & $97.2\%$ & $\textbf{0.94}$ \\
\hline
\end{tabular}
}
\end{footnotesize} 
\label{tab:plogp_ablation4}
\end{table}
\begin{table}[!ht]
\centering
\begin{footnotesize}
\subfloat[\textbf{Similarity threshold $\delta=0.6$}]
{
\begin{tabular}{ccc:ccc:c}
\multicolumn{3}{c}{\textit{Training configuration}} & \multicolumn{3}{c:}{\textit{Generation task}} & \textit{Regression}\\
Model & Numerical Encoding & Self-consistency ($\alpha$) & \textbf{Improvement} & Similarity $\delta$  & Success rate  & Pearson's $r$ (PCC) \\
\hline
JT-VAE & -- & -- & $0.21_{\pm0.7}$ & $\textbf{0.69}_{\pm0.1}$ & $46.4\%$ & \textit{Unfeasible} \\
GCPN & -- & -- & $0.79_{\pm0.6}$ & $0.68_{\pm0.1}$ & $\textbf{100}\%$ & \textit{Unfeasible}\\
\textbf{RT} & \checkmark & $1$ &  $\textbf{2.21}_{\pm1.3}$ & $\textbf{0.69}_{\pm0.1}$ & $81.7\%$ & $0.92$ \\
\textbf{RT} & \cmark & $0$ &  $2.04_{\pm1.3}$ & $\textbf{0.69}_{\pm0.1}$ & $75.0\%$ & $0.90$ \\
\textbf{RT} & \xmark & $1$ &  $2.10_{\pm1.3}$ & $\textbf{0.69}_{\pm0.1}$ & $81.1\%$ & $0.91$ \\
\textbf{RT} & \xmark & $0$ &  $2.10_{\pm1.3}$ & $\textbf{0.69}_{\pm0.1}$ & $81.6\%$ & $\textbf{0.94}$ \\
\hline
\end{tabular}
}
\end{footnotesize} 
\label{tab:plogp_ablation6}
\end{table}
\FloatBarrier
\newpage
\subsection{Protein sequence language modeling}
\subsubsection{Impact of loss functions (Protein interaction data)}
\label{sec:ablation_boman}
Like for the QED dataset, for protein sequence modeling we also investigated the impact of the three training loss setups:
\begin{enumerate}[noitemsep,topsep=0pt]
    \item the vanilla PLM objective~(\autoref{eq:plm}),
    \item alternating objective with the PLM-based text loss~($\mathcal{J}_{G}$;~\autoref{eq:cg_objective}, equivalent to setting $\alpha=0$ in~\autoref{eq:sc_objective}),
    \item alternating objectives with the self-consistency term~($\mathcal{J}_{SC}$;~\autoref{eq:sc_objective}; $\alpha=1$).
\end{enumerate}
The results in~\autoref{tab:bomanablation} show that the proposed training scheme with alternating optimization of property tokens and text tokens was highly effective for both, the regression and the generation task.
\begin{table}[!htb]
\vspace{-4mm}
\centering
\captionsetup{font=footnotesize}
\captionof{table}{
\textbf{Ablation study on training schemes for Boman dataset.}
Legend like Table~\ref{tab:plmqed}.
}
\begin{footnotesize}
\begin{tabular}{cc:cc:cc}
& & \multicolumn{2}{:c:}{\textit{Regression task}} & \multicolumn{2}{c}{\textit{Generation task}} \\
Model & Loss & RMSE ($\downarrow$) & Pearson's $r$ ($\uparrow$) & $0$-Var ($\downarrow$) & Spearman $\rho$ ($\uparrow$) \\
\cline{1-6}
$k$-NN & -- & $0.53$ & $0.932$ & \multicolumn{2}{c}{\textit{Task unfeasible}} \\
RT & PLM & $0.69_{\pm0.03}$ & $0.944_{\pm0.02}$ & $0.3_{\pm0.4}$ & $0.76_{\pm0.03}$ \\
RT & $\mathcal{J}_G$ & $\textbf{0.17}_{\pm0.04}$ & $\textbf{0.994}_{\pm0.01}$ & $\textbf{0.2}_{\pm0.1}$ & $0.82_{\pm0.01}$  \\
RT & $\mathcal{J}_{SC}$ & $0.20_{\pm0.04}$ & $0.991_{\pm0.01}$ & $\textbf{0.2}_{\pm0.1}$ & $\textbf{0.84}_{\pm0.00}$  \\
\cline{1-6}
\end{tabular}
\end{footnotesize} 
\vspace{-2mm}
\label{tab:bomanablation}
\end{table}
In addition, like on the QED dataset, the self-consistency loss led to better results in conditional generation, but at the expense of slightly reduced accuracy in regression.
As stated in the main text, this is most likely caused by the self-evaluations of the decoded sequences. 
These sequences might differ significantly from the training sequences but are still used with the property value of the original sequences. 
Since the Boman index can be computed directly from the sequence, this hypothesis could, in principle, be confirmed by correcting the property value during the self-evaluation call.
However, limited value would come from such approach because real datasets work with more complex properties.

\begin{figure}[!htb]
\centering
\includegraphics[width=0.4\linewidth]{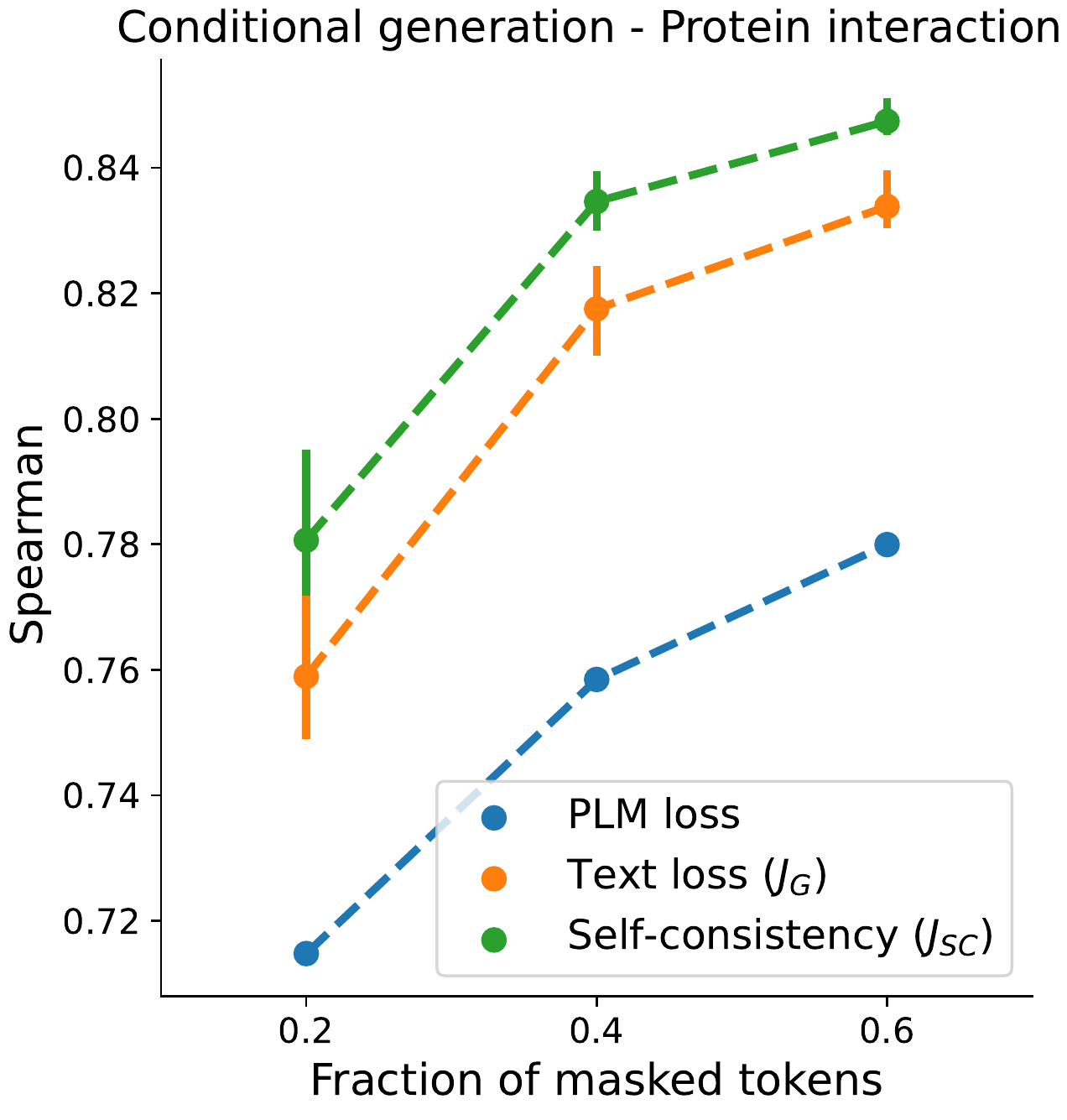}
\caption{
\begin{small}
\textbf{Correlation between property primer and property of generated protein sequences}
The model's ability to generate protein sequences with a desired protein interaction index.
The self-consistency loss yielded the best results and, generally, a higher fraction of masked tokens led to generated peptides that adhere better to the primed property value.
Note that the Boman/protein interaction index can be assessed \textit{in-silico} from the sequence alone.
\end{small}
}
\label{fig:boman_ablation}
\end{figure}

Apart from that,~\autoref{fig:boman_ablation} reveals a general trend in the conditional generation with the Regression Transformer: More freedom in the generative process (i.e., a higher fraction of masked amino acid residues) leads to better results in terms of Spearman $\rho$ to the property primers (cf.~\autoref{fig:boman_ablation}).
This comes, however, at the cost of reduced similarity to the seed sequence.

\FloatBarrier
\subsection{Chemical reaction modeling}
This subsection lists additional results related to the reaction yield modeling.
\autoref{tab:buchwald_additive} reports an ablation study on the impact of $p_{mask}$ (i.e., the probability to mask a specific token) for the reconstruction of additives in Buchwald-Hartwig aminations. 
\begin{table}[!htb]
    \centering
\begin{tabular}{ ccc }
$p_{mask}$ & Top-3 acc. & Tanimoto sim. \\ \hline
$1.0$ & $1.36\%_{\pm{0.5}}$ & $0.158_{\pm{0.002}}$ \\ 
$0.5$  & $11.47\%_{\pm{1.0}}$ & $0.316_{\pm{0.002}}$ \\ 
$0.25$  & $46.74\%_{\pm{3.5}}$ & $0.645_{\pm{0.003}}$ \\ \hline
\end{tabular}
    \vspace{2mm}
    \caption{Performance in generating additives for Buchwald-Hartwig reactions~\citep{ahneman2018predicting} as a function of $p_{mask}$, i.e., the fraction of tokens in the additive that are masked. Generation was primed with remaining precursors and yield.
}
\label{tab:buchwald_additive}
\end{table}
\autoref{tab:yield_buch_extended} and~\autoref{tab:yield_suz_extended} report an ablation study that assess whether co-encoding reaction yield enables the model to better reconstruct precursors.
\begin{table}[!htb]
\centering
\begin{tabular}{ c  | cc|  cc|  c | c  }
Precursor & \multicolumn{2}{c|}{Top-3 accuracy} &
\multicolumn{2}{c|}{Tanimoto similarity} & \multirow{2}{*}{Unique $n$}& \\
  type &  Prec. + Yield & Precursors  & Prec. + Yield  & Precursors &   \\ \hline
Aryl halide & $98.23\%_{\pm{0.5}}$ & $98.21\%_{\pm{0.4}}$ & $0.991_{\pm{0.003}}$ & $0.991_{\pm{0.002}}$  & 15\\ 
Ligand & $50.38\%_{\pm{1.6}}$ & $50.43\%_{\pm{1.7}}$ &  $0.677_{\pm{0.010}}$ & $0.678_{\pm{0.010}}$ &  4\\ 
Base & $100.0\%_{\pm{0.0}}$ & $100.0\%_{\pm{0.6}}$ & $1.000_{\pm{0.000}}$  &  $1.000_{\pm{0.000}}$ & 3\\ 
Additive & $1.36\%_{\pm{0.5}}$ & $1.25\%_{\pm{0.8}}$ & $0.158_{\pm{0.018}}$ & $0.158_{\pm{0.019}}$ & 22\\ \hline
\end{tabular}
\vspace{2mm}
\captionof{table}{
Generating precursors for Buchwald-Hartwig reactions~\citep{ahneman2018predicting} based on remaining precursors or remaining precursors and yield.
Full precursors were generated ($p_{mask}=1$).
Unique $n$ denotes the number of unique samples per entity in the training dataset.
}
\label{tab:yield_buch_extended}
\end{table}


\begin{table}[!htb]
\centering
\begin{tabular}{ c  | cc|  cc|  c  }
Precursor & \multicolumn{2}{c|}{Top-3 accuracy} &
\multicolumn{2}{c|}{Tanimoto similarity} &  \multirow{2}{*}{Unique $n$} \\
  type &  Prec. + Yield & Precursors  & Prec. + Yield  & Precursors &  \\ \hline
Electrophile & $44.19\%_{\pm{17.6}}$ & $31.39\%_{\pm{15.3}}$ & $0.732_{\pm{0.160}}$ & $0.591_{\pm{0.141}}$  & 7\\ 
Nucleophile & $100.0\%_{\pm{0.0}}$ & $100.0\%_{\pm{0.0}}$ &  $1.000_{\pm{0.000}}$ & $1.000_{\pm{0.000}}$  & 4\\ 
Ligand & $67.43\%_{\pm{20.0}}$ & $67.59\%_{\pm{19.8}}$ & $0.689_{\pm{0.152}}$  &  $0.690_{\pm{0.152}}$ & 5\\ 
Base & $90.53\%_{\pm{1.2}}$ & $90.50\%_{\pm{1.4}}$ & $0.811_{\pm{0.006}}$ & $0.811_{\pm{0.001}}$  & 8\\
Solvent & $56.74\%_{\pm{1.1}}$ & $56.52\%_{\pm{1.0}}$ & $0.661_{\pm{0.009}}$ & $0.660_{\pm{0.007}}$ & 4\\ \hline
\end{tabular}
\vspace{2mm}
\captionof{table}{
Generating precursors for Suzuki-cross-couplings reactions~\citep{perera2018platform} based on remaining precursors or remaining precursors and yield.
Full precursors were generated ($p_{mask}=1$).
Unique $n$ denotes the number of unique samples per entity in the training dataset.
}
\label{tab:yield_suz_extended}
\end{table}

\begin{figure}
    \centering
    \includegraphics[width=1.0\linewidth]{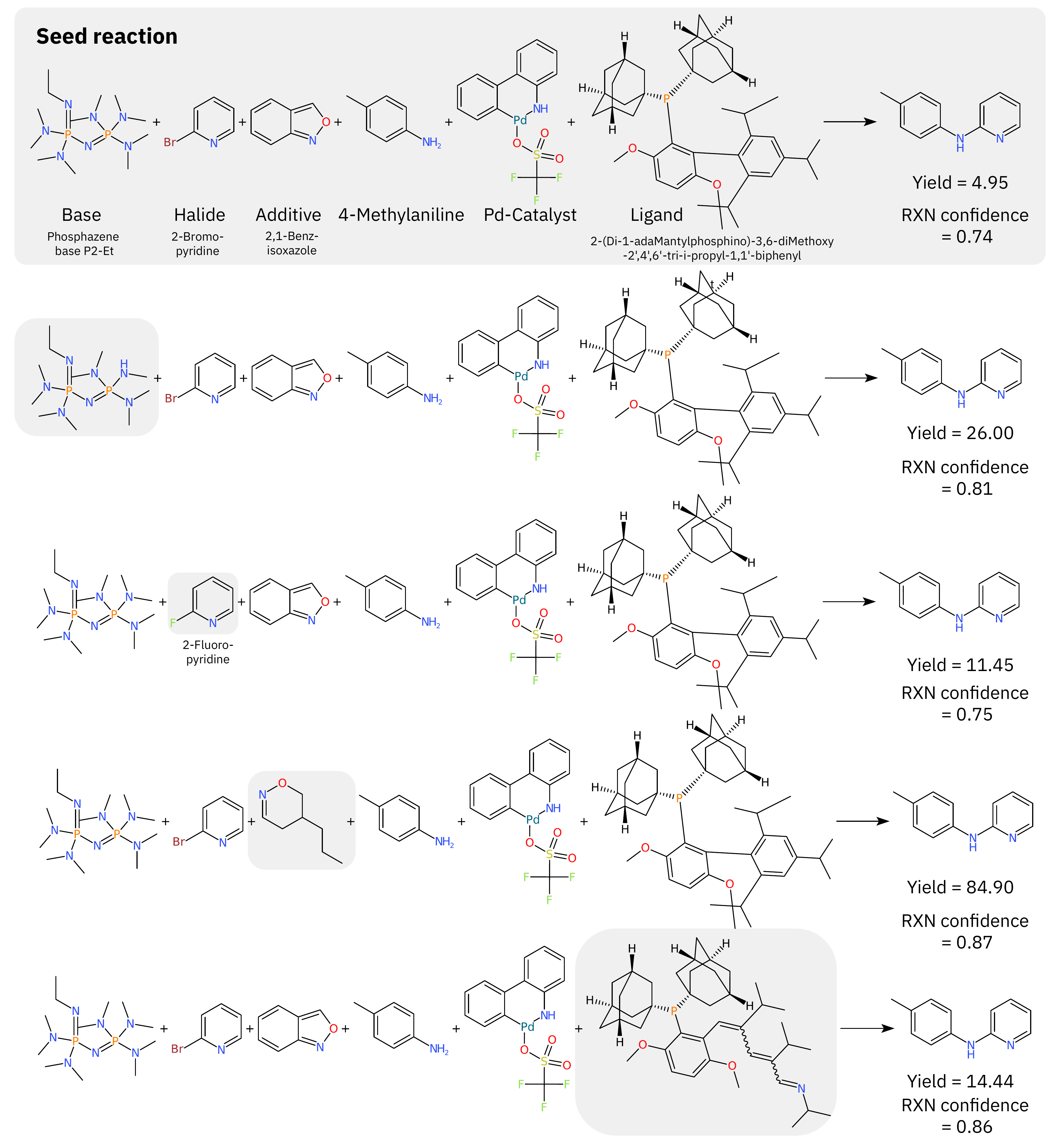}
    \caption{
    Adapting an unseen Buchwald-Hartwig amination toward higher yield.
    Alongside a seed reaction and its reported yield, the RT can generate reactions that selective replace individual precursors. 
    In this case, upon priming for higher yield and a given precursor type, the RT indeed generated reactions with higher yield (as predicted by the RT) as well as higher confidence for the reaction to suceed in general (predicted with forward model from~\citep{schwaller2019molecular}).
    Note that no adaptations of 4-Methylaniline and the Palladium-catalyst are generated since they are constanta cross the dataset.
    This is the full figure of~\autoref{fig:yield_example_main} (main manuscript)
    }
    \label{fig:yield_example_full}
\end{figure}

\FloatBarrier
\section{Case studies}
\subsection{Case study on scaffold hopping}
\label{sec:scaff}
Scaffold hopping is a technique in medicinal chemistry with the goal to discover novel compounds by modifying the central core structure (i.e., removing substituents while retaining rings and their linker fragments) of known compounds~\citep{bajorath2017computational}. 
We simulated this task on the QED dataset by determining the scaffold with~\texttt{RDKit} and masking only the non-scaffold tokens (in contrast to the regular evaluation where randomly $40\%$ of the tokens were masked).
This task was only performed with the SMILES models since scaffolds cannot be determined trivially in SELFIES. 
In general, this task is more challenging because the molecule is more constrained. 
On average, less tokens are being masked and in most cases the full range of drug-likeness cannot be captured, given the scaffold.
This explains the higher percentage of molecules where the primer did not influence the generations (cf. Table~\ref{tab:qed_scaff}).
\begin{table}[!htb]
\centering
\captionsetup{font=footnotesize}
\captionof{table}{
\textbf{Scaffold hopping performance for SMILES model.} 
No numerical encodings were used. 
No standard deviations are available for the scaffold results since the masking is deterministic.
}
\begin{small}
\begin{tabular}{ccccc}
Text loss & Task & $0$-Var ($\downarrow$) & Spearman's $\rho$ ($\uparrow$) \\
\cline{1-5}
$\mathcal{J}_G$ & Masking non-scaffold & $8.55\%$ & $0.136$\\
$\mathcal{J}_{SC}$ & Masking non-scaffold  & $9.76\%$ & $0.105$\\
$\mathcal{J}_G$ & Masking randomly & $0.80\%_{\pm0.19}$ & $0.108_{\pm0.01}$\\
$\mathcal{J}_{SC}$ & Masking randomly & $1.14\%_{\pm0.19}$ & $0.085_{\pm0.02}$\\
\cline{1-5}
\end{tabular}
\end{small} 
\label{tab:qed_scaff}
\end{table}

\begin{figure*}[!htb]
\centering
\includegraphics[width=0.8\linewidth]{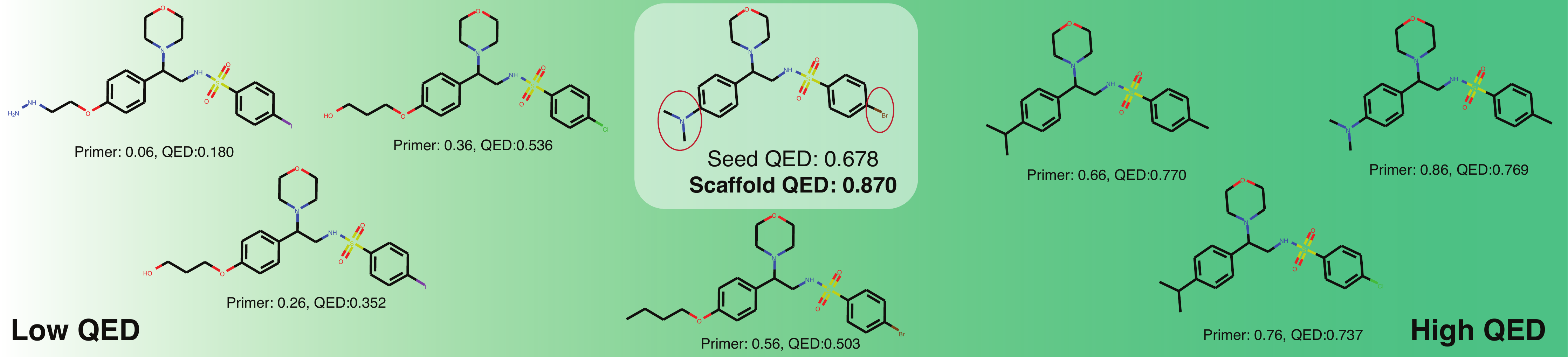}
\caption{
\begin{small}
\textbf{Molecules sampled in a scaffold hopping task.}
Only non-scaffold tokens (encircled in red) were masked.
\end{small}
}
\label{fig:scaffoldmols}
\end{figure*}
But note, that this includes cases where the molecule is itself a scaffold and thus no tokens are masked (we do not control for that explicitly).
The generations for one exemplary molecule are shown in Figure~\ref{fig:scaffoldmols}. 
In this example, it is interesting to see that the model decorated the scaffold with specific atoms on the rightmost six-ring.
These atoms, iodine, chlorine and bromine which were rightfully provided from low to high QED primers seem to be indicative of different levels of drug-likeness.
One drawback, however, is that the RT cannot fill no or multiple tokens in the position of one~\texttt{[MASK]} location. 
For example, in the case of the last primer ($0.86$), the provided scaffold already had a QED of $0.87$ and thus not adding any new atoms would have been the best choice here.

\newpage
\subsection{Case study on attention visualization: Interpreting attention heads}
\label{sec:attention}
\begin{figure*}[!htb]
\centering
\includegraphics[width=1.0\linewidth]{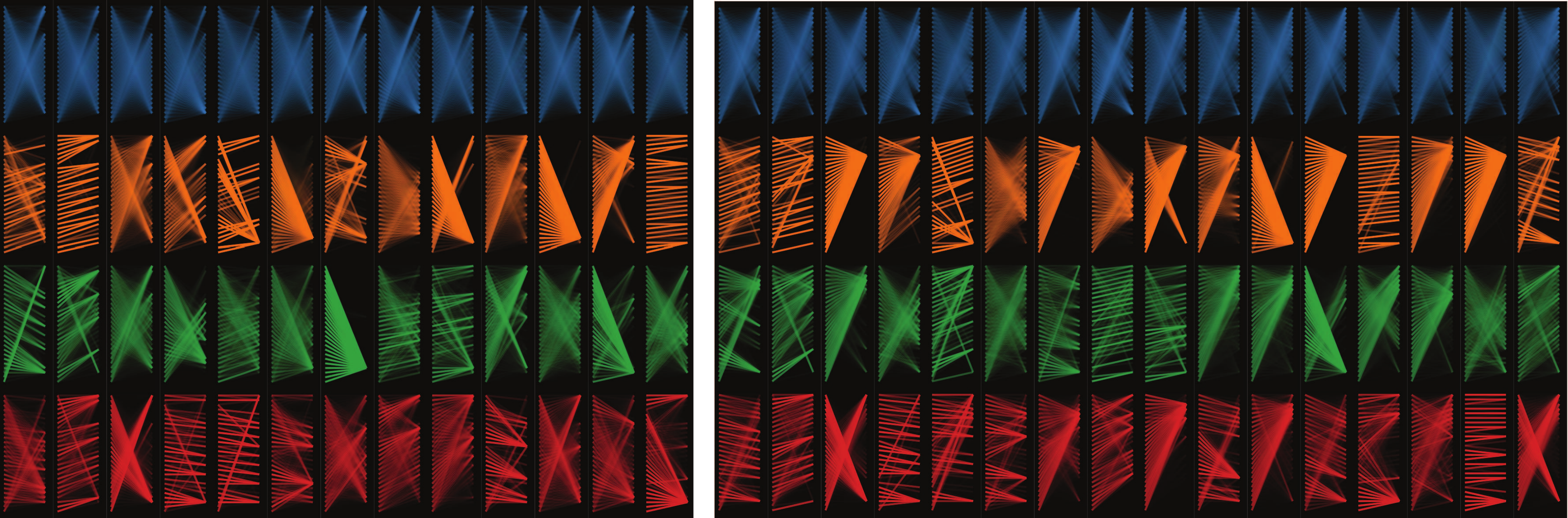}
\caption{
\begin{small}
\textbf{
Comparing attention scores across both tasks with BertViz~\citep{vig2019multiscale}.
}
Attention scores for all heads of the first four layers. Rows depict layers, column depict attention heads. Within each cell, the tokens are ordered from top to bottom.
\textit{Left:} Property prediction task.
\textit{Right:} Conditional generation task.
Plot performed with SELFIES model with float encodings, trained on the self-consistency loss.
\end{small}
}
\label{fig:att}
\end{figure*}

We visualized the attention scores using BertViz~\citep{vig2019multiscale}.
Here, we aimed to compare the inference patterns across the two tasks, property prediction and conditional generation.
The results for the first $4$ (out of $32$) layers are shown in Figure~\ref{fig:att}.
In general, many attention patterns commonly described in natural language models are also present in the Regression Transformer. 
For example the bag-of-words pattern (i.e., evenly distributed attention, e.g., all heads of first layer) or the next-token (e.g., layer $4$, head $4$ and $5$) or previous-token patterns (e.g., layer $2$, head $2$) are clearly visible. 
While the named patterns are consistently present in both tasks, probably because they are useful irrespective of the particular task, some distinctive patterns for either of the tasks can be found.
For example, in the conditional generation task (Figure~\ref{fig:att}, right) many triangles with their right angle in the upper right are present. 
In these positions the property tokens are present and thus these patterns indicate that the representation of all other tokens, especially also the masked ones, are heavily influenced by the property value.
Instead, in the property prediction task (Figure~\ref{fig:att}, left), many triangles with their right angle in the lower right are present. 
This implies a heavy attention on the~\texttt{[END]} token which marks the end of the sequence and is a useful indicator for the QED score because it is critically influenced by the size/weight of the molecule.
One particularly interesting attention head is head $3$ in layer $2$. 
In the property prediction task its role is to make the masked property tokens aware of the sequence length. 
In the conditional generation task, its role is to make all tokens aware of the property values.

\section{Additional results}
\subsection{Failure mode in fluourescence dataset from TAPE}
\label{sec:failure}
This dataset is particularly interesting due to its bimodal mode: one mode corresponding to bright proteins, the other to dark proteins (cf.~\autoref{fig:tapeplot}).
An interesting failure mode was observed when initially training on the Fluorescence dataset. 
\autoref{fig:tapeplot} shows that the dark mode has one sharp spike, exactly at a log fluorescence value of $1.301$. 
Almost $10\%$ of all training samples and almost $50\%$ of the proteins in the dark mode have this exact value.
The Regression Transformer is trained on a classification loss and so, the loss during training for such samples will be distributed across the five tokens \texttt{1\_0}, \texttt{.}, \texttt{3\_0}, \texttt{0\_-1} and \texttt{1\_-2}.
In many cases, the model collapsed to always predicting $3.301$ where the first token (\texttt{3\_1}) was correct for all samples in the bright mode and the remaining tokens (\texttt{3\_0}, \texttt{0\_-1} and \texttt{1\_-2} were correct for most samples in the dark model.
This happened because no weighting of the individual numerical tokens was applied.
As a non-algorithmic remedy, we added Gaussian noise to those training samples.

\subsection{Detailed performance - Classification vs. Regression}
\label{sec:fluor}
\begin{figure}[!htb]
\centering
\includegraphics[width=0.7\linewidth]{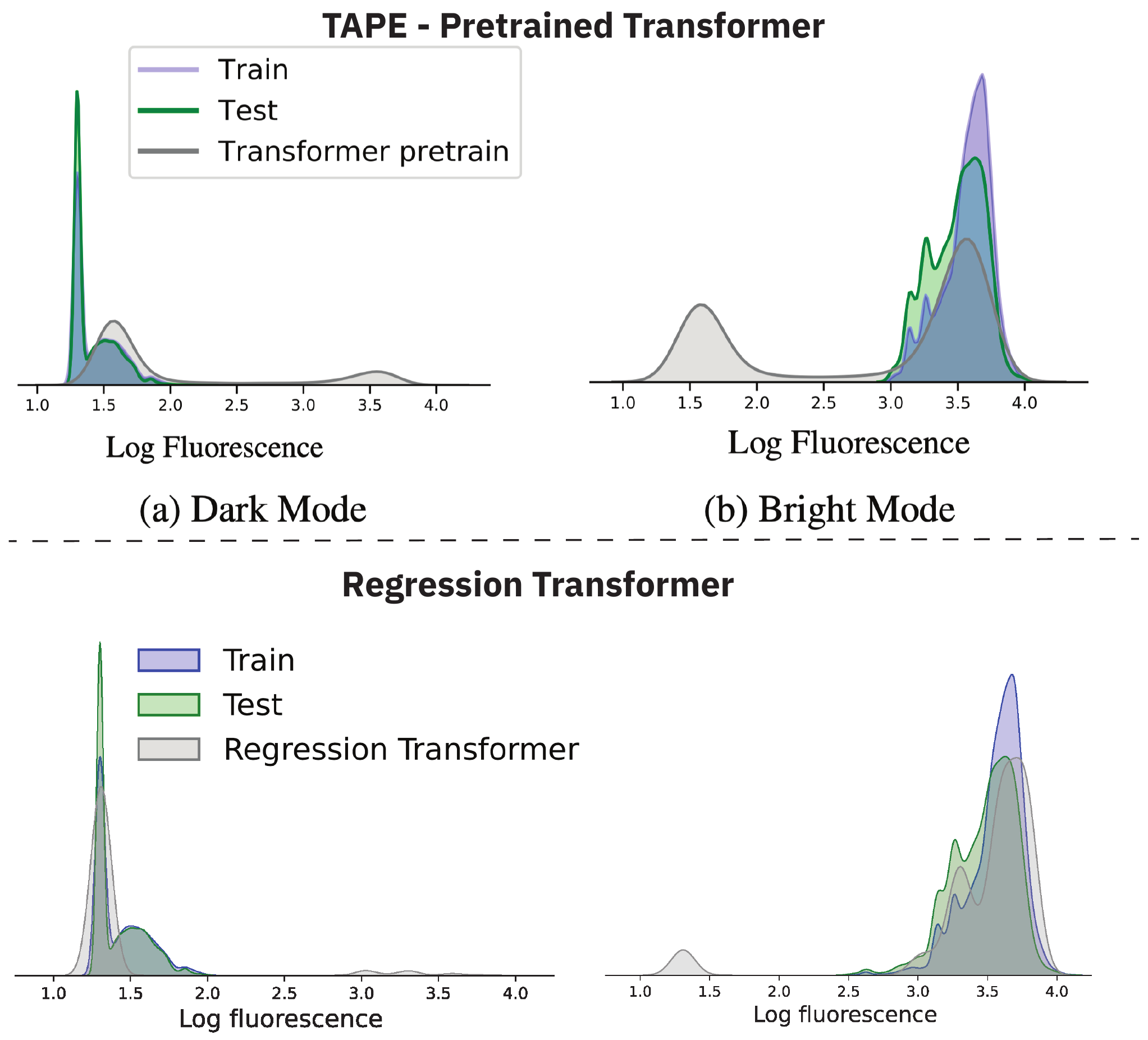}
\caption{
\begin{small}
\textbf{Bimodal mode of fluorescence data.}
The upper part of the plot has been copied from~\citet{rao2019evaluating} (Figure 3). 
It shows the bimodal mode of the training data and the test predictions from the TAPE Transformer.
At the bottom, we show our remake of the above plot by replacing the predictions from the pretrained TAPE Transformer with the predictions from the Regression Transformer.
\end{small}
}
\label{fig:tapeplot}
\end{figure}

\autoref{fig:tapeplot} reveals the improved performance of the RT compared to the finetuned TAPE Transformer: Less samples were predicted in the wrong mode. 
However, the RT had difficulties with a fine grained regression, in particular in the bright mode.
This becomes particularly apparent when inspecting the detailed results, grouped by bright and dark test proteins respectively (cf.~\autoref{tab:fluorregression}).
\begin{table}[!htb]
\centering
\captionsetup{font=footnotesize}
\captionof{table}{
\textbf{Detailed fluorescence prediction results.}
MSE abbreviates mean squared error. 
TAPE and UniRep performances taken from~\citet{rao2019evaluating}.
For the RT all standard deviations on $\rho$ and MSE were $<0.05$ and $<0.1$ respectively.
}
\begin{footnotesize}
\begin{tabular}{cccccccc}
&  & \multicolumn{2}{c}{\underline{Full test set}} & \multicolumn{2}{c}{\underline{Bright proteins}} & \multicolumn{2}{c}{\underline{Dark proteins}} \\
Model & Source & MSE & $\rho$ & MSE & $\rho$ & MSE & $\rho$ \\
\cline{1-8}
One-Hot & TAPE & $2.69$ & $0.14$ & $0.08$ & $0.03$ & $3.95$ & $0.00$ \\
$k$-NN & \textbf{Ours} & $2.31$ & $0.59$ & $\textbf{0.05}$ & $0.30$ & $3.37$ & $0.04$ \\
Pretr. LSTM & TAPE & $\textbf{0.19}$ & $0.67$ & $0.12$ & $0.62$ & \textbf{0.22} & 0.04 \\
Pretr. Transf. & TAPE & 0.22 & $0.68$ & $0.09$ & 0.60 & 0.29 & \textbf{0.05} \\
UniRep & UniRep & 0.20 & $0.67$ & 0.13 & \textbf{0.63} & 0.24 & 0.04 \\
\textbf{RT} & \textbf{Ours} & 0.34 & \textbf{0.72} & 0.19 & 0.45 & 0.40 & 0.04  \\
\cline{1-8}
\end{tabular}
\end{footnotesize} 
\label{tab:fluorregression}
\end{table}
While the RT achieved the best results in the overall Spearman $\rho$, the recommended metric by~\cite{rao2019evaluating}, it does not dominate any of the mode-specific metrics. 
This is a noteworthy finding because it reflects the tendency of the RT to strive for a multi-class classification rather than performing a full regression.
It is also interesting to see that the baseline models ($k$-NN and TAPE One-Hot) achieved the best results in MSE of bright proteins.

\FloatBarrier

\end{document}